\documentclass{bmvc2k}
\usepackage{booktabs}
\usepackage{pifont}
\usepackage{multirow}
\usepackage{amsfonts}

\title{Outlier detection by ensembling uncertainty with negative objectness}

\addauthor{Anja Delić}{anja.delic@fer.hr}{1}
\addauthor{Matej Grcić}{matej.grcic@fer.hr}{1}
\addauthor{Siniša Šegvić}{sinisa.segvic@fer.hr}{1}

\addinstitution{
University of Zagreb,\\
Faculty of Electrical Engineering and Computing, \\
Zagreb, Croatia
}

\runninghead{Delić, Grcić, Šegvić
}{Ensembling uncertainty with negative objectness}

\begin{document}
\newcommand{\added}[1]{{{\textcolor{blue}{#1}}}}
\maketitle

\begin{abstract}
Outlier detection is an essential capability
of safety-critical visual recognition.
Many existing methods deliver good results
by encouraging standard closed-set models
to produce low-confidence predictions 
in negative training data.
However, that approach conflates
prediction uncertainty with
recognition of outliers.
We disentangle the two factors
by revisiting the K+1-way classifier
that involves K known classes
and one negative class.
This setup allows us
to formulate a novel outlier score
as an ensemble of in-distribution uncertainty 
and the posterior of the negative class
that we term negative objectness.
Our UNO score can detect outliers
due to either high prediction uncertainty
or similarity with negative training data.
We showcase the utility of our method 
in experimental setups with 
K+1-way image classification
and K+2-way dense prediction.
In both cases we show that the bias
of real negative data can be relaxed
by leveraging a 
jointly trained 
normalizing flow.
Our models outperform the current state-of-the art
on standard benchmarks for image-wide
and pixel-level outlier detection.

\end{abstract}

\section{Introduction}
\label{introduction}

Modern machine learning 
\cite{he2016deep,dosovitskiy21iclr} delivers unprecedented performance on a plethora of academic datasets \cite{deng09cvpr,lin14eccv,cordts2016cityscapes}.
These 
evaluation protocols are often limited to instances of predetermined taxonomies \cite{cordts2016cityscapes,neuhold2017mapillary}.
However, many objects in the real world 
do not belong to the training taxonomy.
In such situations, 
standard algorithms for image recognition \cite{liu2022cvpr}
and scene understanding \cite{chen18tpami,cheng2022masked}
may 
behave unpredictably \cite{zendel18eccv}.



Discriminative models 
may become robust to outliers
through upgrade to 
open-set recognition 
\cite{liang22neurips}.
This can be conveniently achieved
by complementing standard classification
with outlier detection \cite{zhang20eccv}.
Typically, the outlier detector
delivers a scalar outlier score
that 
induces ranking and enables detection through thresholding \cite{ruff21pieee}.
Many outlier detectors
both in the image-wide
\cite{hendrycks2018deep, dhamija18neurips}
and the pixel-level context
\cite{bevandic22ivc, biase2021cvpr, 
  chan2021entropy, grcic22eccv, tian2022pixel}
rely on negative training data. 
Although it cannot represent 
the entire variety of the visual world,
the negative data can still help by signaling
that not all data should be confidently recognized.
If the variety of the negative data greatly
exceeds the variety of the inliers,
then there is a reasonable hope
that the test outliers will be detected.
Unfortunately, this entails 
undesired bias towards test outliers 
that appear similar to the negative training data.
This concern can be addressed
either by relying on synthetic negatives 
\cite{lee18iclr,neal18eccv, grcic24tpami},
or through separate ranking with respect to approaches
that do not train on real negative data 
\cite{zhang2023openood,blum2021fishyscapes,chan2021segmentmeifyoucan}.
\begin{figure}[ht]
    \centering
    \includegraphics[width=\textwidth]{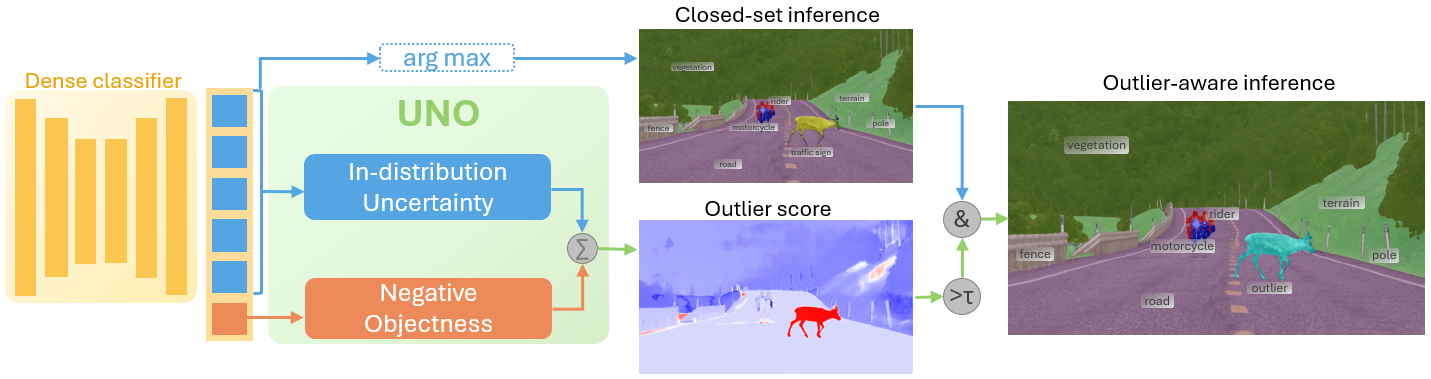}
    \caption{We propose UNO, a plug-in module that enables outlier-aware inference atop a desired feature extractor. 
    UNO decouples in-distribution uncertainty from outlier recognition, 
    and boosts the outlier detection performance by ensembling the two components.}
    \label{fig:uno_overview}
\end{figure}

Many existing methods 
improve open-set performance 
of  existing closed-set models 
by encouraging
low-confidence predictions in negative training data \cite{hendrycks19icml,lee18iclr}.
However, this approach conflates prediction uncertainty with outlier recognition.
In this paper, we disentangle the two factors by reconsidering the K+1-way classifier where the additional class 
represents the negative data.
This allows us 
to formulate the outlier score 
as the posterior of the K+1-th class,
which we term negative objectness.
Moreover, the outlier score 
can also be formulated
as prediction uncertainty 
over the K known classes \cite{hendrycks19icml}.
We find that these two formulations 
exhibit complementary behaviour 
due to being sensitive to different test outliers.
Consequently, we propose a novel outlier score
that we term UNO:
ensemble of \emph{\textbf{U}ncertainty} over K inlier classes \cite{hendrycks17iclr} and \emph{\textbf{N}egative \textbf{O}bjectness}.


Our UNO score is a strikingly good fit 
in the pixel-level context
as an extension of some
direct set-prediction approach \cite{carion2020eccv}.
We express the negative objectness 
as the posterior of the negative class 
in the frame of a K+2-way mask-level classifier \cite{cheng2022masked,yu22eccv,li23cvpr}.
A closer look suggests that our score performs well
due to different inductive bias 
of the two components
as indicated by a remarkably weak correlation.
The UNO score performs competitively
in the image-wide context as well,
even though the absence 
of the no-object class
decreases the synergy of the two components. 
Our models outperform the state-of-the-art
on several standard benchmarks
in both learning setups,
\textit{i.e.}~with and without real negative data.

\section{Related work}

\paragraph{Image-wide setup.}
Image-wide classification models
can be complemented with outlier detection
either through a modified training procedure
or in a post-hoc manner with frozen parameters \cite{zhang2023openood}.
Post-hoc methods build upon pre-trained closed-set classifiers.
Early baselines express the outlier score as
maximum softmax probability \cite{hendrycks2016baseline} or maximum logit \cite{hendrycks2019scaling}. 
TempScale \cite{guo2017calibration} calibrates softmax probabilities with temperature scaling while ODIN \cite{liang2017enhancing} pre-processes the input with anti-adversarial perturbations. 
Recent post-hoc methods  simplify layer activations \cite{djurisic2022extremely} or 
fit generative models to pre-logit features 
\cite{zhang2022out, yu22eccv, taghikah24aaai}.
In contrast, training methods involve various regularizations that either 
consider only inlier data or
utilize real negative data as well. 
The former rely on modeling confidence \cite{devries2018arxiv},
self-supervised training \cite{hendrycks2019using, tack2020neurips}, per-sample temperature scaling \cite{hsu2020cvpr},
outlier synthesis in feature space \cite{du2022iclr, tao2023iclr, kumar23cvpr} or mitigating overconfidence \cite{wei2022icml}.
Recent methods guide the learning process with respect to nearest neighbours \cite{park2023iccvb} or leverage distances in feature space either by combining different distance functions \cite{noh2023iccv} or distances in different feature spaces \cite{chen2023iccv}. 
Early work in training with negative data \cite{hendrycks2018deep, dhamija18neurips} encourages high entropy in inlier samples. 
MCD \cite{yu2019unsupervised} promotes discrepancy between prediction on OOD samples of two classification heads.
UDG \cite{yang2021semantically} groups negative data following inlier taxonomy to enrich semantic knowledge of inlier classes.
MixOE \cite{zhang2023mixture} interpolates between inliers and negatives to improve regularization.
Unlike related training-based methods, UNO does not modify the standard classifier training objective, but
merely requires a slight architectural change 
to extend the classifier with a logit for the negative class. Moreover, empirical insights suggest that UNO requires significantly less negative data than related methods.

\paragraph{Dense prediction.}
Many image-wide outlier detection methods can directly complement models for pixel-level prediction \cite{hendrycks2016baseline, hendrycks2019scaling}.
Early approaches estimate the prediction uncertainty with maximum 
softmax probability \cite{hendrycks17iclr},
ensembling \cite{lakshminarayanan2017simple} or Bayesian uncertainty \cite{mukhoti2018evaluating}. 
Subsequent work suggests
energy scores \cite{liu2020neurips}
and the standardized max-logit score \cite{jung2021standardized}.
Training with negative data can be incorporated by pasting negative content from broad negative datasets (ImageNet \cite{bevandic19gcpr}, ADE20K \cite{zhou2017scene} or COCO \cite{lin2014microsoft}) atop inlier images
\cite{biase2021cvpr, chan2021entropy, grcic22eccv, hendrycks2018deep, tian2022pixel, bevandic22ivc}.  
However, real negative data can be replaced with synthetic negative patches obtained by sampling a jointly trained generative model \cite{lee18iclr, neal18eccv, zhao21arxiv}.
Outlier detection performance can be further improved by providing more capacity \cite{vaze2021open}.
Another 
branch of methods 
considers
generative models \cite{blum2021fishyscapes, grcic22eccv, liang22neurips}. 

\noindent
\paragraph{Mask-level recognition.}
Recent panoptic architectures
\cite{cheng2022masked,yu22eccv,li23cvpr}
decompose scene understanding 
into class-agnostic segmentation
and region-wide recognition.
They frame the detection of semantic masks
as direct set-prediction 
where each mask is classified into
K inlier classes and one no-object class.
This induces a degree of openness 
and therefore favours outlier detection. 
Recent work shows 
great applicability of the Mask2Former architecture 
for open-set segmentation 
\cite{nayal2023rba, rai2023unmasking, ackermann2023maskomaly, grcic2023advantages}.
All these methods detect anomalies 
due to not belonging to any inlier class 
and improve when training with negative data.
Specifically, RbA \cite{nayal2023rba} 
reduces energy in negative pixels, 
Mask2Anomaly \cite{rai2023unmasking} 
uses a contrastive loss, 
while EAM \cite{grcic2023advantages} 
ensembles region-wide outlier scores.
However, these approaches 
conflate negative data
with no-object regions.  
Furthermore, they fail to exploit the fact 
that mask queries behave like 
one-vs-all classifiers 
\cite{nayal2023rba, ackermann2023maskomaly},
which makes them appropriate 
for direct detection of anomalous regions.
UNO addresses these weaknesses 
by decoupling negatives 
from the no-object class,
and representing them
as a standalone class 
in the K+2-way taxonomy. 

\section{Disentangling negative objectness from uncertainty}
\label{method}
Let
$h_{\theta_1}: \mathcal{X} \rightarrow \mathbb{R}^d$ be a feature extractor 
(e.g.\ ResNet \cite{he2016deep} or 
 ViT \cite{dosovitskiy21iclr})
 onto which 
we attach our K+1-way classifier 
$g_{\theta_2}: 
 \mathbb{R}^d \rightarrow \Delta^{K}$, 
where 
$\mathcal{X}$ is the input space and
$\Delta^{K}$ is a K-dimensional 
probabilistic simplex.
Given an input 
$\mathbf{x} \in \mathcal{X}$, 
we compute a latent representation 
$\mathbf{z} = h_{\theta_1}(\mathbf{x})$ 
and map it onto the simplex point 
$\mathbf{p} = 
  \text{softmax} (g_{\theta_2}(\mathbf{z}))$.
Thus, our deep model 
$f_\theta = g_{\theta_2} \circ h_{\theta_1}$
maps input samples from $\mathcal{X}$ 
onto simplex points in $\Delta^K$.
We assume that the classifier $g_{\theta_2}$
is a simple projection.
Thus we have $g_{\theta_2}(\textbf{x}) = \textbf{W}\cdot\textbf{x}$,
where the rows of $\textbf{W}$ correspond
to class vectors $\textbf{w}_j$.


\noindent
\paragraph{Negative objectness.}
Given feature representation $\mathbf{z}$ 
and a linear classifier $g$, 
we formulate 
the negative objectness score 
($\textbf{s}_\text{NO}$)
as posterior of the K+1-th class:
\begin{equation}
\label{eq:s_no}
    \mathbf{s}_{\text{NO}}(\mathbf{z}) := P(Y=\text{K+1} | \mathbf{z}) 
    = \frac{\text{exp}(\mathbf{w}_{K+1}^T \mathbf{z})}{\sum_{j=1}^{K+1} \text{exp}(\mathbf{w}_j^T \mathbf{z})}
    \;. 
\end{equation}
The $\textbf{s}_\text{NO}$ score tests whether a given sample is semantically dissimilar from the inlier classes and similar to the negative training data.
Consequently, a test outlier 
that is similar to the training negatives 
will yield high $\textbf{s}_\text{NO}$.

\noindent
\paragraph{Prediction uncertainty.}
Negative objectness is not suitable
for detecting test outliers 
that are semantically dissimilar 
from the negative training data.
However, many of these outliers yield 
uncertain predictions across inlier classes 
of our K+1-classifier.
We address this observation 
by defining the uncertainty score
as negative prediction confidence
across the K inlier classes:
\begin{equation}
\label{eq:s_unc}
    \mathbf{s}_\text{Unc}(\mathbf{z}) := - \max_{k=1...K} P(Y=k|\mathbf{z})  
    = - \max_{k=1...K} 
      \frac{\text{exp}(\mathbf{w}_k\mathbf{z})}
         {\sum_{j=1}^{K+1} \text{exp}(\mathbf{w}_j\mathbf{z})} 
    = - 
      \frac{\text{exp}(\mathbf{w}_{\hat{k}} \mathbf{z})}
        {\sum_{j=1}^{K+1} \text{exp}(\mathbf{w}_j \mathbf{z})} 
    \;. 
\end{equation}
The last equality in (\ref{eq:s_unc}) denotes the index of the winning class as $\hat{k} = \text{argmax}_k \, \exp(\mathbf{w}_k^T \mathbf{z})$.
The uncertainty score $\textbf{s}_\text{Unc}$ 
detects different outliers 
than negative objectness $\textbf{s}_\text{NO}$ 
despite attending to 
the same shared features 
$\mathbf{z}$.


\noindent
\paragraph{UNO score.}
We integrate the 
prediction uncertainty (\ref{eq:s_unc})
and negative objectness (\ref{eq:s_no}) 
into the UNO score as the sum of the two terms: 
\begin{equation}
       \mathbf{s}_{\text{UNO}}(\mathbf{z})
    = \mathbf{s}_{\text{Unc}}(\mathbf{z}) + \mathbf{s}_{\text{NO}}(\mathbf{z}).
\end{equation}

UNO has 
an interesting geometrical interpretation 
in the pre-logit space $\mathbb{R}^d$.
Class vectors 
tend to be mutually orthogonal:
$\mathbf{w}_i^T \mathbf{w}_j = 0, \,\, \forall i,j \in \{1, 2, ..., K+1\}$ \cite{wang2022vim}.
This arrangement is enforced
by the standard supervised loss ($- \log \text{softmax}$) 
whenever there is enough data,
and 
enough capacity and dimensionality ($d >K+1$)
in the feature extractor $h_{\theta_1}$.
Thus, a small angle between 
the latent representation $\textbf{z}$ 
and the negative vector $\textbf{w}_{K+1}$ 
leads to high outlier objectness $\textbf{s}_\text{NO}$.
On the other hand,
the uncertainty score $\textbf{s}_\text{Unc}$
is negatively correlated 
with the feature norm $\lvert\lvert \mathbf{z} \rvert\rvert$.
If we fix the direction of $\textbf{z}$,
its norm can be viewed 
as reciprocal softmax temperature.
As usual,
large temperatures
lead to uncertain predictions \cite{hinton2015arxiv}. 

\begin{figure}[ht]
    \centering
    \includegraphics[width=\linewidth]{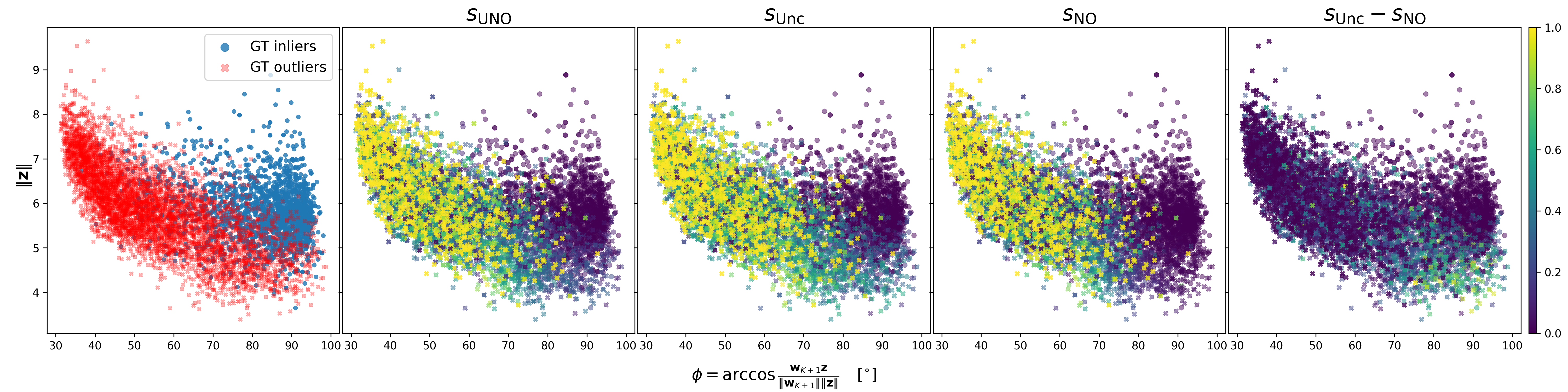}
    \caption{
    Interpretation of UNO in the pre-logit space $\mathbb{R}^d$ in an OpenOOD \cite{zhang2023openood} experiment  
    with CIFAR-10 inliers.
    Small $L_2$ norm of feature representations $\mathbf{z}$ leads to high $\textbf{s}_\text{Unc}$ while the angle between $\mathbf{z}$ and the K+1-th weight vector $\mathbf{w}_\text{K+1}$ leads to high $\textbf{s}_\text{NO}$. 
    The two components 
    capture different outliers as shown on the rightmost plot.
    }
    \label{fig:angle_vs_norm}
\end{figure}

The above insights show 
that the two components of our score
will be at least partially decorrelated
due to being sensitive to different outliers:
$\textbf{s}_\text{NO}$ detects large feature norms 
and good alignment with $\mathbf{w}_{K+1}$, 
while $\textbf{s}_\text{Unc}$ detects small feature norms
and poor alignment with inlier vectors.
Thus, the UNO score is likely
to achieve lower error
than any of the two components alone.
Figure \ref{fig:angle_vs_norm} illustrates
our insight 
in an image-wide recognition experiment on small images.
Further empirical analysis indicates a weak positive correlation between the UNO components, as detailed in the Appendix.

%

\noindent
\paragraph{Extending UNO for dense prediction.} 
The UNO outlier detector can be attached to any pre-trained deep classifier.
In the case of image-wide prediction, we simply extend a pre-trained K-way classifier with an additional class.
However, in the case of dense prediction, many contemporary architectures rely on mask-level recognition \cite{cheng2022masked}.
Consequently, we build our per-pixel score by averaging the mask-level UNO scores:
\begin{align}
    \label{eq:uno_m2f}
   \textbf{s}_{\text{UNO}}^{\text{M2F}} = 
   \sum_{i=1}^N \mathbf{m}_i  \cdot
      \textbf{s}_{\text{UNO}}(\textbf{z}_i)
      \;. 
\end{align}
Note that 
equation~\ref{eq:uno_m2f} assumes that the dense prediction model 
has partitioned the input image into $N$ different masks $\mathbf{m}_i$ with the corresponding latent 
mask embeddings $\mathbf{z}_i$.

We note that the mask-level classifiers
already use the K+1-th logit 
to indicate that the particular query
has not been associated 
with any image region.
It would be a bad idea to reuse that logit 
for mask-level negative objectness
since unused queries are regular 
occurrences in in-distribution images.   
Consequently, we extend 
the default classifier
with the negative class.
In our implementations,
the K+1-th logit 
corresponds to negative objectness
while the K+2-th logit 
indicates the no-object class.
Please find more details in the Appendix.

\section{Training UNO with and without real negative data}

Let 
$\mathcal{D}_\text{in} = 
  \{(\textbf{x}_i, y_i)\}_{i=1}^{N^+}$ 
be an inlier dataset
with $\mathbf{x} \in \mathcal{X}$ and 
$y_i \in \mathcal{Y}$, 
where $\mathcal{Y}$ is 
the set of classes of size $K=|\mathcal{Y}|$.
Let $\mathcal{D}_\text{out} = 
  \{\textbf{x}_j\}_{j=1}^{N^-}$  
be a negative dataset that mimics test outliers. 
We construct a mixed-content training set $\mathcal{D}$ 
as the union of inlier samples with negative samples labeled as the K+1-th class: $\mathcal{D} =  \mathcal{D}_\text{in} \cup \{(\mathbf{x}_j, K+1) \, | \, \textbf{x}_j \in \mathcal{D}_\text{out}\}$.

We start from a K-way classifier 
that is pre-trained on inlier data  
$D_\text{in}$ 
as a part of an off-the-shelf model.
We append the K+1-th class 
and fine-tune the classifier $f_\theta$ 
on the mixed-content dataset $\mathcal{D}$
by optimizing the standard cross-entropy loss:
\begin{equation}
    L_\text{cls}(\theta) = \mathbb E_{\mathbf{x}, {y} \in \mathcal{D}} [- \ln P (y | \mathbf{x})]\, .
    \label{eq:cls}
\end{equation}
Every training minibatch contains 
an equal number of samples for all K+1 classes. 
Thus, we expose our model to significantly less negative data than the alternative approaches
\cite{zhang2023mixture, yu2019unsupervised, yang2021semantically, hendrycks17iclr}, which utilize $2\times$ more negatives than inliers.
Even though the outlier exposure 
of our model is weak, 
it will still be biased 
towards detection of the test outliers 
that are similar to the training negatives.
This issue can be circumvented by replacing
the auxiliary negative dataset 
with synthetic negatives 
produced by a generative model
\cite{lee18iclr,grcic21visapp}.

\noindent
\paragraph{Synthetic negatives.}
We generate model-specific synthetic negatives 
by a normalizing flow $f_\psi$
\cite{dinh17iclr,kingma2018neurips}
that we train alongside 
our discriminative classifier $f_\theta$.
We train the flow to generate samples
that resemble inliers, but at the same time
give raise to uncertain predictions 
across the inlier classes
according to the following loss 
\cite{lee18iclr,grcic24sensor}: 
\begin{equation}
  L_{\text{flow}}(\psi, \theta) = 
    L_{\text{mle}}(\psi) + 
    \beta \cdot L_{\text{jsd}}(\psi, \theta). 
  \label{eq:nf-joint}
\end{equation}
The first loss term 
$L_\mathrm{mle}(\psi) = 
  \textbf{E}_{\mathbf{x^+} \in D_\text{in}}
    [ - \ln p_\psi(\mathbf{x})]$
maximizes the likelihood of inlier samples,
while the second loss term
$L_{\text{jsd}}(\psi, \theta) = 
  \text{JSD}(U, f_\theta(\mathbf{x}))$
minimizes the 
Jensen-Shannon
divergence 
between the uniform distribution 
and the predictions over the K inlier classes.
The two competing objectives settle down 
when the generative model produces samples 
at the border of the inlier manifold 
\cite{lee18iclr}.
The above loss
jointly optimizes
$f_\theta$ and $f_\psi$
so that the discriminative model 
get aware of the negative class
and thus preclude feature collapse
\cite{lucas19neurips}.
Of course, the training also has to ensure
that $f_\theta$ satisfies 
the primary discriminative loss. 

However, a naive application 
of this recipe in the K+1-way context
would not work as intended.
In fact, the generative loss (\ref{eq:nf-joint})
and the discriminative loss (\ref{eq:cls})
would simply be satisfied 
by confident classification
of all generated samples
into the negative class.
This trivial solution 
would embed the generated samples
into a compact region
of the feature space,
and make them useless
as proxies for outlier detection.
Thus, we propose 
a two-step optimization procedure.
The first step jointly trains 
$f_\psi$, $f_\theta$, 
and the K inlier logits 
according to (\ref{eq:nf-joint}).
The second step freezes the flow
and fine-tunes the loss (\ref{eq:cls}) 
in order to ensure recognition 
of generated negatives as the K+1-th class.
Different from previous approaches \cite{lee18iclr,grcic24sensor}, 
the loss (\ref{eq:nf-joint}) 
serves only to promote 
generation of useful synthetic negatives, 
while we ensure the outlier recognition quality 
by subsequent fine-tuning 
according to the loss (\ref{eq:cls}).

\begin{figure}[ht]
    \centering
    \includegraphics[width=\textwidth]{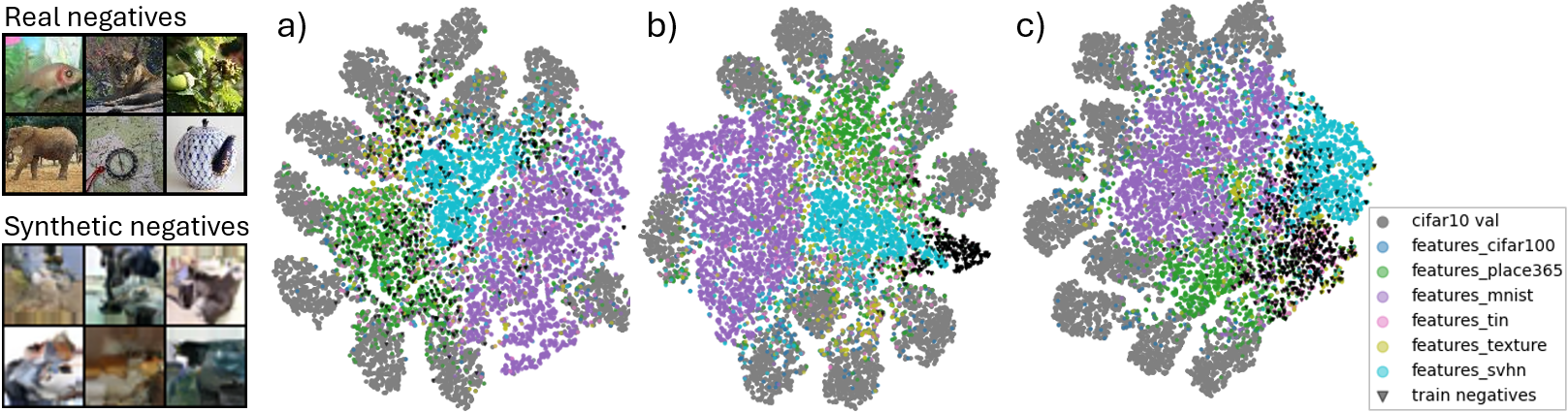}
    \caption{
    Left: Visualization of real and synthetic negatives from an OpenOOD CIFAR-10 
    experiment \cite{zhang2023openood}. 
    Right: t-SNE plots 
    of the corresponding feature representations.
    Our two-set training strategy 
    yields synthetic negatives 
    near the inlier manifold (a), 
    while the naive approach 
    collapses synthetic negatives 
    to a single mode (b).
    Relative location of real negatives 
    indicate that they cover 
    similar modes of test outliers 
    as our synthetic samples (c).
    }
    \label{fig:scatter}
\end{figure}

Figure \ref{fig:scatter}.a illustrates 
the learned latent representations 
as obtained by the two-step training procedure.
Our synthetic negatives are sprinkled
around the inlier manifold.
Joint optimisation of the loss (\ref{eq:nf-joint}) 
and the K+1-way cross-entropy loss (\ref{eq:cls}) 
collapses synthetic negatives to a single mode, 
as shown in Figure \ref{fig:scatter}.b.
For reference, we also show the latent representations 
of the model trained with real negatives 
in Figure \ref{fig:scatter}.c.

\begin{figure}[ht]
    \centering
    \includegraphics[width=0.88\textwidth]{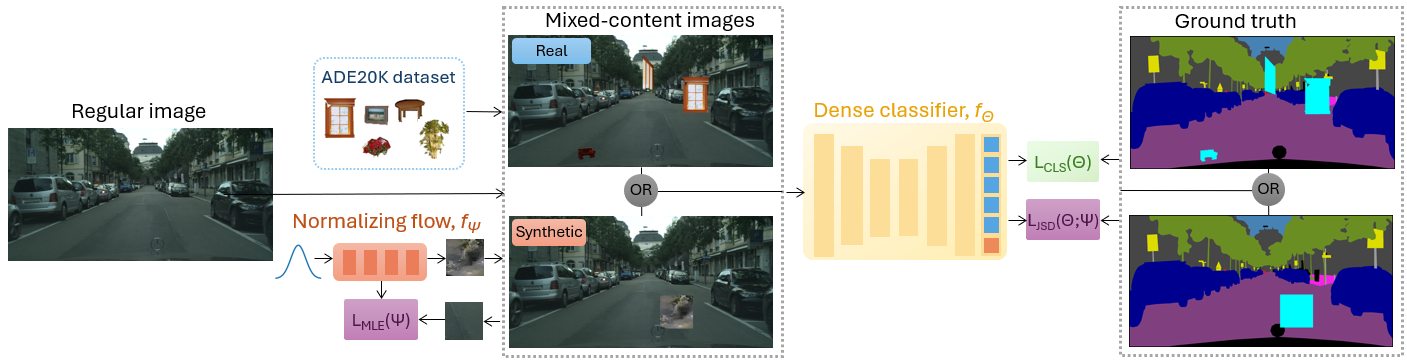}
    \caption{
    Fine-tuning of dense classifier equipped with UNO. 
    We paste negative training data (either real or synthetic) atop regular inlier images. The resulting mixed-content image is fed to the dense classifier that optimizes cross-entropy loss over K+2 classes.
    }
    \label{fig:uno_training}
\end{figure}

Figure \ref{fig:uno_training} shows 
our training approach in the dense prediction context.
We paste negative training content atop the regular image to produce a mixed-content image.
The mixed-content image is fed to the K+2-way classifier that optimizes the cross-entropy objective, similar as in the image-wide context.
In the case of synthetic negatives, we jointly optimize the dense classifier and the normalizing flow.

\section{Experiments}
\label{sec:experiments}
\paragraph{Experimental setup.}
We evaluate the UNO performance on standard image-wide and pixel-level benchmarks.
In the pixel-level setup, we consider Fishyscapes \cite{blum2021fishyscapes} and SMIYC \cite{chan2021segmentmeifyoucan}, the two prominent benchmarks for road driving scenes.
In the image-wide setup, we consider the CIFAR-10 and ImageNet-200 setups from the OpenOOD benchmark \cite{zhang2023openood}.
We report the standard evaluation metrics AUROC and FPR\textsubscript{95} in image-wide,
and AP and FPR\textsubscript{95} in pixel-level experiments.
Our pixel-level experiments start by training the M2F-SwinL \cite{cheng2022masked} on Cityscapes \cite{cordts2016cityscapes} and Vistas \cite{neuhold2017mapillary} with the Cityscapes taxonomy.
We extend the mask-wide classifier to K+2 classes and fine-tune on mixed-content images. 
We compose mixed-content images 
by pasting negative content
(ADE20k \cite{zhou2019semantic} instances or flow samples)
into inlier images. 
More implementation details are in the Appendix. Our code is publicly available here\footnote{\url{https://github.com/matejgrcic/Open-set-M2F}}.

\noindent
\paragraph{Segmentation of road scenes.}
Table~\ref{tab:fishyscapes_smiyc} 
compares UNO with the related work on Fishyscapes and SMIYC.
We observe that mask-level recognition methods outperform
earlier works \cite{jung2021standardized,grcic22eccv,vojir2021road}.
When training with real negatives, our method outperforms all previous work on Fishyscapes and SMIYC AnomalyTrack, while performing within variance of the best method on SMIYC ObstacleTrack.
When training without real negatives, our method outperforms all previous work on SMIYC and Fishyscapes Static by a large margin, while achieving second best AP on Fishyscapes L\&F.
Interestingly, the inlier recognition performance is not affected by the UNO fine-tuning.
The initial closed-set model attains 83.5\% mIoU on Cityscapes.
Appending UNO and fine-tuning the resulting open-set model with real negatives increases classification performance to 83.7\%.

\begin{table}[ht]
\setlength{\tabcolsep}{2pt}
\small
\centering
  \footnotesize
  \begin{tabular}{lcccccccccc}
    & & \multicolumn{4}{c}{Fishyscapes} & \multicolumn{4}{c}{SMIYC} & Cityscapes \\
    & Aux. & \multicolumn{2}{c}{Lost\&Found} & \multicolumn{2}{c}{Static} 
     & \multicolumn{2}{c}{AnomalyTrack} & \multicolumn{2}{c}{ObstacleTrack}
     & val \\
    Method & data & AP & $\text{FPR}_{95}$ & AP & $\text{FPR}_{95}$ & AP & $\text{FPR}_{95}$ & AP & $\text{FPR}_{95}$ & mIoU\\
    \midrule
    Maximum Entropy \cite{chan2021entropy} & \ding{55} & 15.0 & 85.1 & 0.8 & 77.9 & - & - & - & -& 9.7 \\
    Image Resynthesis \cite{lis2019detecting} & \ding{55} & 5.7 & 48.1 & 29.6 & 27.1 & 52.3 & 25.9 & 37.7 & 4.7 & 81.4 \\
    JSRNet \cite{vojir2021road} & \ding{55} & - & - & - & - & 33.6 & 43.9 & 28.1 & 28.9 & - \\
    Max softmax \cite{hendrycks2016baseline} & \ding{55} & 1.8 & 44.9 & 12.9 & 39.8 & 28.0 & 72.1 & 15.7 & 16.6 & 80.3 \\
    SML \cite{jung2021standardized} & \ding{55} & 31.7 & 21.9 & 52.1 & 20.5 & - & - & - & - & -\\
    Embedding Density \cite{blum2021fishyscapes} & \ding{55} & 37.5 & 70.8 & 0.8 & 46.4 & 4.3 & 47.2 & 62.1 & 17.4 & 80.3 \\
    NFlowJS \cite{grcic21visapp} & \ding{55} & 39.4 & 9.0 & 52.1 & 15.4 & - & - & - & - & 77.4 \\
    SynDHybrid \cite{grcic23arxiv} & \ding{55} & 51.8 & 11.5 & 54.7 & 15.5 & - & - & - & - & 79.9 \\
    cDNP \cite{galesso23iccv} & \ding{55} & \textbf{62.2} & \textbf{8.9} & - & - & 88.9 & 11.4 & 72.70 & 1.40 & - \\
    $\text{EAM}^\dagger$ \cite{grcic2023advantages} & \ding{55} & 9.4 & 41.5 & 76.0 & 10.1 & 76.3 & 93.9 & 66.9 & 17.9 & 83.5 \\
    $\text{RbA}^\dagger$ \cite{nayal2023rba} & \ding{55} & - & - & - & - & 86.1 & 15.9 & 87.8 & 3.3 & - \\
    $\text{Maskomaly}^\dagger$ \cite{ackermann2023maskomaly} & \ding{55} & - & - & - & - & 93.4 & 6.9 & - & - & - \\
    $\text{UNO}^\dagger$ (ours) & \ding{55} & \underline{56.4} & 55.1 & \textbf{91.1} & \textbf{1.5} & \textbf{96.1} & \textbf{2.3} & \textbf{89.0} & \textbf{0.6} & 83.5 \\
    \midrule
    SynBoost \cite{di2021pixel} & \ding{51} & 43.2 & 15.8 & 72.6 & 18.8 & 56.4 & 61.9 & 71.3 & 3.2 & 81.4 \\
    OOD Head \cite{bevandic22ivc} & \ding{51} & 30.9 & 22.2 & 84.0 & 10.3 & - & - & - & - & 77.3 \\
    Void Classifier \cite{blum2021fishyscapes} & \ding{51} & 10.3 & 22.1 & 45.0 & 19.4 & 36.6 & 63.5 & 10.4 & 41.5 & 70.4 \\
    Dirichlet prior \cite{malinin2018predictive} & \ding{51} & 34.3 & 47.4 & 84.6 & 30.0 & - & - & - & - & 70.5 \\
    DenseHybrid \cite{grcic22eccv} & \ding{51} & 43.9 & 6.2 & 72.3 & 5.5 & 78.0 & 9.8 & 78.7 & 2.1 & 81.0 \\
    PEBAL \cite{tian2022pixel} & \ding{51} & 44.2 & 7.6 & 92.4 & 1.7 & 49.1 & 40.8 & 5.0 & 12.7 & - \\
    cDNP \cite{galesso23iccv} & \ding{51} & 69.8 & 7.5 & - & - & 88.9 & 11.4 & 72.70 & 1.40 & - \\
    RPL \cite{liu2023iccv} & \ding{51} & 53.9 & 2.3 & 95.9 & 0.5 & 83.5 & 11.7 & 85.9 & 0.6 & -\\
    $\text{Mask2Anomaly}^\dagger$ \cite{rai2023unmasking} & \ding{51} & 46.0 & 4.4 & 95.2 & 0.8 & 88.7 & 14.6 & \textbf{93.3} & 0.2 & -\\
    $\text{RbA}^\dagger$ \cite{nayal2023rba} & \ding{51} & - & - & - & - & 90.9 & 11.6 & 91.8 & 0.5 & -\\
    $\text{EAM}^\dagger$ \cite{grcic2023advantages} & \ding{51} & 63.5 & 39.2 & 93.6 & 1.2 & 93.8& 4.1 & 92.9 & 0.5 & 83.5 \\
    $\text{UNO}^\dagger$ (ours) & \ding{51} & \textbf{74.8} & \textbf{2.7} & \textbf{95.8} & \textbf{0.3} & \textbf{96.3} & \textbf{2.0} & \underline{93.2} & \textbf{0.2} & 83.7 \\
\end{tabular}
\vspace{0.2cm}
    \caption{Experimental evaluation on the Fishyscapes and SMIYC benchmarks. Methods that leverage mask-level recognition are marked with  $\dagger$. Missing results are marked with -.}
    \label{tab:fishyscapes_smiyc}
\end{table}

Table~\ref{tab:fs-val} compares UNO with previous works that work with mask-level recognition on validation subsets of Fishyscapes and RoadAnomaly \cite{lis2019detecting}.
UNO outperforms previous approaches across datasets and metrics with the single exception of 0.5 pp worse $\text{FPR}_{95}$ than RbA on RoadAnomaly. 
\begin{table}[!htb]
\setlength{\tabcolsep}{4pt}
\centering
\small
    \footnotesize
  \begin{tabular}{lcccccc|cccccccc}
     & \multicolumn{2}{c}{FS L\&F} & \multicolumn{2}{c}{FS Static} & \multicolumn{2}{c|}{RoadAnomaly}
     & \multicolumn{2}{c}{FS L\&F} & \multicolumn{2}{c}{FS Static} & \multicolumn{2}{c}{RoadAnomaly}\\
    Method & AP & $\text{FPR}_{95}$ & AP & $\text{FPR}_{95}$ & AP & $\text{FPR}_{95}$
    & AP & $\text{FPR}_{95}$ & AP & $\text{FPR}_{95}$ & AP & $\text{FPR}_{95}$\\
    \midrule
    RbA & 
        70.8 & 6.3 & - & - & 85.4 & \textbf{6.9} &
        61.0 & 10.6 & - & - & 78.5 & 11.8\\
    Maskomaly &
        69.4 & 9.4 & 90.5 & 2.0 & 79.7 & 13.5 
        & - & - & 68.8 & 15.0 & 80.8 & 12.0 \\
    EAM & 
        81.5 & 4.2 & 96.0 & 0.3 & 69.4 & 7.7 &
        52.0 & 20.5 & 87.3 & 2.1 & 66.7 & 13.4\\
    UNO (ours) & 
        \textbf{81.8} & \textbf{1.3} & \textbf{98.0} & \textbf{0.04} & \textbf{88.5} & \underline{7.4} &
        \textbf{74.5} & \textbf{6.9} & \textbf{96.9} & \textbf{0.1} & \textbf{82.4} & \textbf{9.2}\\
\end{tabular}
\vspace{0.2cm}
  \caption{Validation of mask-level approaches on Fishyscapes val and RoadAnomaly with (left) and without (right) training with real negative data. Missing results are marked with -.}
    \label{tab:fs-val}
\end{table}

\noindent
\paragraph{Image classification.}
\label{subsec:exp_image_level}
Table\ \ref{tab:open-ood} shows the image-wide performance of UNO on the OpenOOD benchmark \cite{zhang2023openood}.
We compare UNO with training methods with and without the use of real negative data. 
When training with real negatives, UNO consistently outperforms all baselines on CIFAR-10 and near OOD Imagenet-200, while attaining the best AUROC and the second-best FPR on the far OOD Imagenet-200.
When training without real negative data, UNO accomplishes competitive performance on both benchmarks.
Synthetic UNO outperforms even methods trained with real data, specifically, MixOE, MCD and UDG on CIFAR-10, and UDG on ImageNet-200. 
It is interesting that UNO trained with synthetic negatives outperforms UNO trained with real negatives by a significant margin on the far OOD ImageNet-200.
Finally, UNO preserves the inlier classification performance.
\begin{table}[ht]
\centering
  \footnotesize

  \begin{tabular}{l
   c@{\ }c@{\quad}c@{\ }c@{\ }c@{\quad\quad}
   c@{\ }c@{\quad}c@{\ }c@{\ }c@{\quad\quad}
  }
    \multirow{3}{*}{Method} & \multicolumn{5}{c}{CIFAR-10} & \multicolumn{5}{c}{ImageNet-200} \\
    &  \multicolumn{2}{c}{Near-OOD} & \multicolumn{2}{c}{Far-OOD} && \multicolumn{2}{c}{Near-OOD} & \multicolumn{2}{c}{Far-OOD} &  \\

     & AUC & FPR & AUC & FPR & Acc. 
     & AUC & FPR & AUC & FPR & Acc.\\
    \midrule
    ConfBranch \cite{devries2018arxiv}
        & 89.8 & 31.3 & 92.9 &94.9 & 94.9
        & 79.1 & 61.4 & 90.4 & 34.8 & 85.9\\
    RotPred \cite{hendrycks2019using}
        & \textbf{92.7} & \textbf{28.1} & \textbf{96.6} & \textbf{12.2} & 95.4
        & 81.6 & 60.4 & 92.6 & 26.2 & 86.4 \\
    G-ODIN \cite{hsu2020cvpr}
        & 89.1 & 45.5 & 95.5 & 21.5 & 94.7 
        & 77.3 & 69.9 & 92.3 & 30.2& 84.6\\
    CSI \cite{tack2020neurips}
        & 89.5 & 33.7 & 92.0 & 26.4 & 91.2 
        & - & - & - & - & -  \\
    ARPL \cite{chen2022pami}
        & 87.4 & 40.3 & 89.3 & 32.4 & 93.7
        & 82.0 & \textbf{55.7} & 89.2 & 36.5 & 84.0  \\
    MOS \cite{huang2021cvpr}
        & 71.5 & 78.7 & 76.4 & 62.9 & 94.8
        & 69.8 & 71.6 & 80.5 & 51.6 & 85.6  \\
    VOS \cite{du2022iclr}
        & 87.7 & 57.0 & 90.8 & 40.4 & 94.3
        & 82.5 & 59.9 & 91.0 & 34.0 & 86.2 \\
    LogitNorm \cite{wei2022icml}
        & 92.3 & 29.3 & 96.7 & 13.8 & 94.3
        & \textbf{82.7} & 56.5 & 93.0 & 26.1 & 86.0 \\
    CIDER \cite{ming2023iclr}
        & 90.7 & 32.1 & 94.7 & 20.7 & - 
        & 80.6 & 60.1 & 90.7 & 30.2 & -  \\
    NPOS \cite{tao2023iclr}
        & 89.8 & 32.6 & 94.1 & 20.6 & -
        & 79.4 & 62.1 & \textbf{94.5} & \textbf{21.8} & - \\
    UNO (ours) 
        & 91.3 & 31.8 & 92.6 & 20.5 & 95.2  
        & 81.2 & 61.1 & 92.5 & 32.3 & 86.3 \\
    
    \midrule
    MixOE \cite{zhang2023mixture}
        & 88.7 & 51.5 & 91.9 & 33.8 & 94.6
        & 82.6 & 58.0 & 88.3 & 40.9 & 85.7 
    \\
    MCD \cite{yu2019unsupervised}
        & 91.0 & 30.2 & 91.0 & 32.0 & 95.0 
        & 83.6 & 54.7 & 88.9 
        & \textbf{29.9} & 86.1 \\
    UDG \cite{yang2021semantically}
        & 89.9
        & 35.3
        & 92.4
        & 20.4
        & 92.4
        & 74.3
        & 68.9
        & 82.1
        & 62.0
        & 68.1
    \\
    OE \cite{hendrycks17iclr}
        & 94.8 & 19.8 & 96.0 & 13.1& 94.6 
        & 84.8 & 52.3 & 89.0 & 34.2 & 85.8 
        \\
    UNO (ours)
        & \textbf{94.9} & \textbf{9.3} 
        & \textbf{97.6} & \textbf{9.4} 
        & 94.9
        & \textbf{85.1} & \textbf{51.7} 
        & \textbf{89.6} & 36.8 
        & 86.4\\
  \end{tabular}
  \vspace{0.2cm}
  \caption{OOD detection performance on OpenOOD CIFAR-10 and ImageNet-200. 
  We consider training-based methods with (bottom) and without (top) real negative data.
  All results were averaged over 3 runs. Full results are available in the Appendix.
  }
    \label{tab:open-ood}
\end{table}

\noindent
\paragraph{Validating components of UNO score.}
\label{subsec:ablation}
Table \ref{tab:ensemle-ablation} validates the contribution of the UNO components.
In the case of real negatives, both components attain competitive results while UNO significantly outperforms both of them.
In the case of synthetic negatives, 
uncertainty outperforms the negative objectness by a wide margin.
This does not come as a surprise since the negative objectness is inferred from synthetic training negatives.
Still, UNO again greatly benefits from the ensemble of the two scores.
\begin{table}[ht]
\setlength{\tabcolsep}{4pt}
\centering
\small
  \footnotesize
  \begin{tabular}{lcccccc|cccccc}
     \multirow{2}{*}{Score}& \multicolumn{2}{c}{FS L\&F} & \multicolumn{2}{c}{FS Static} & \multicolumn{2}{c|}{RoadAnomaly}
     & \multicolumn{2}{c}{FS L\&F} & \multicolumn{2}{c}{FS Static} & \multicolumn{2}{c}{RoadAnomaly}\\
      & AP  & $\text{FPR}_{95}$ & AP & $\text{FPR}_{95}$ & AP & $\text{FPR}_{95}$
      & AP  & $\text{FPR}_{95}$ & AP & $\text{FPR}_{95}$ & AP & $\text{FPR}_{95}$\\
    \midrule
    UNO 
        & 81.8 &  1.3 & 98.0 &  0.04 & 88.5 &  7.4 
        & 74.5 & 6.9 & 96.9  & 0.1 & 82.4  & 9.2\\
    Unc 
        &74.1 & 4.5   & 72.1 & 1.5     & 66.2 & 8.0
        & 71.9  & 8.0 & 95.7  & 0.5 & 70.4  & 9.4\\
    NO 
        & 69.0 & 1.6   & 92.6 &  0.14    & 80.4 & 19.8
        & 26.6 & 91.1 & 73.9   & 61.1 & 54.2 & 72.3\\
\end{tabular}
\vspace{0.2cm}
  \caption{Validation of UNO components with (left) and without (right) real negative data.}
  \label{tab:ensemle-ablation}
\end{table}

\begin{figure}[ht]
  \centering
  \includegraphics[width=0.95\textwidth]
      {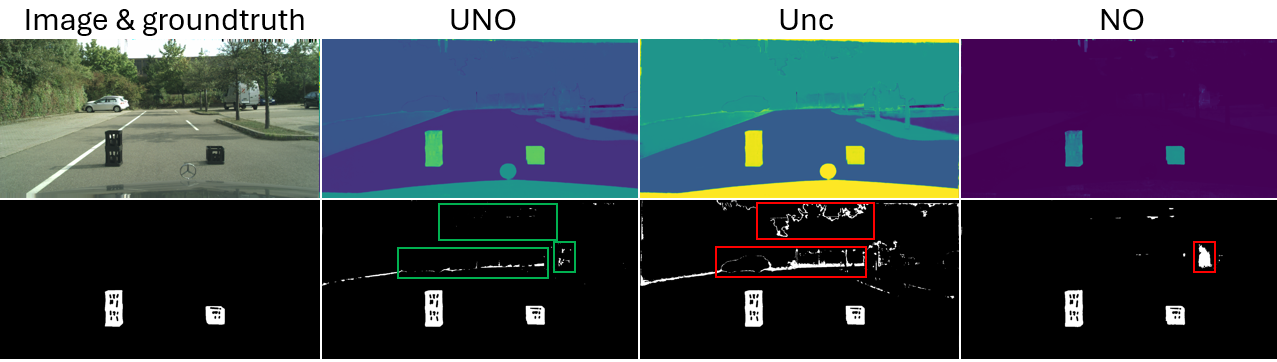} \\
    \caption{Qualitative experiments 
    on Fishyscapes L\&F val. 
    Top row shows the input image and the three outlier scores.
    Bottom row shows anomaly detection maps 
    after thresholding at TPR=95\%. 
    UNO significantly reduces the incidence of false-positive responses.
    }
    \label{fig:examples}
\end{figure}

Figure \ref{fig:examples} visualises UNO performance on FS L\&F. Columns show continuous OOD scores and OOD detections after thresholding at 95\% TPR.
The two UNO components
have different failure modes.
For example, $\textbf{s}_{\text{Unc}}$ produces false positives at the borders of inlier classes while $\textbf{s}_\text{NO}$ detects some inlier objects as outliers.
Still, these failure modes cancel out in the compound UNO score, as designated with green rectangles.

\section{Conclusion}

Visual recognition models behave unpredictably 
in presence of outliers.
A common strategy to detect outliers 
is to fine-tune the closed-set classifier
in order to increase uncertainty in negative data.
We take a different approach and disentangle the prediction uncertainty
from the recognition of outliers. 
This allows us to cast outlier detection
as an ensemble of in-distribution uncertainty
and the posterior of the negative class,
which we term negative objectness.
The resulting UNO score manifests as a lightweight plug-in 
module that can extend an arbitrary classifier.
The resulting model is then fine-tuned with real negative data or their synthetic surrogates.
UNO can be applied to per-pixel outlier detection
by extending recent mask-level recognition approaches.
There we first apply UNO 
to recover mask-wide outlier scores,
and then propagate them to pixels
according to mask assignments.
UNO attains strong performance 
on recent image-wide and pixel-level 
outlier detection benchmarks
with and without real negative data. 
Prominent future developments include 
improving the quality 
of synthetic negative data
and extending our ensemble 
with generative predictions.


\section{Limitations}
\textbf{Auxiliary training data.}
UNO relies on negative training data to deliver strong OOD detection performance.
While standard applications offer plenty of real negative data, this might not be the case in domain-specific applications such as remote sensing and medical imaging.
Still, we show that UNO can deliver strong performance when trained on synthetic negatives.

\noindent
\textbf{Confidence calibration.}
Our K+1-way classifier could be inadequately calibrated which would 
introduce bias into our uncertainty estimates.
An interesting direction for future work is formulation of a better uncertainty estimate. 

\section*{Acknowledgements}
This work has been supported by Croatian Recovery and Resilience Fund - NextGenerationEU (grant C1.4 R5-I2.01.0001), and Croatian Science Foundation (grant IP-2020-02-5851 ADEPT)

\bibliography{bmvc_final}

\begin{thebibliography}{108}
\providecommand{\natexlab}[1]{#1}
\providecommand{\url}[1]{\texttt{#1}}
\expandafter\ifx\csname urlstyle\endcsname\relax
  \providecommand{\doi}[1]{doi: #1}\else
  \providecommand{\doi}{doi: \begingroup \urlstyle{rm}\Url}\fi

\bibitem[Ackermann et~al.(2023)Ackermann, Sakaridis, and Yu]{ackermann2023maskomaly}
Jan Ackermann, Christos Sakaridis, and Fisher Yu.
\newblock Maskomaly: Zero-shot mask anomaly segmentation.
\newblock \emph{arXiv preprint arXiv:2305.16972}, 2023.

\bibitem[Bendale and Boult(2016)]{bendale2016cvpr}
Abhijit Bendale and Terrance~E. Boult.
\newblock Towards open set deep networks.
\newblock In \emph{2016 {IEEE} Conference on Computer Vision and Pattern Recognition, {CVPR} 2016, Las Vegas, NV, USA, June 27-30, 2016}, pages 1563--1572. {IEEE} Computer Society, 2016.

\bibitem[Bevandi{\'c} et~al.(2019)Bevandi{\'c}, Kre{\v{s}}o, Or{\v{s}}i{\'c}, and {\v{S}}egvi{\'c}]{bevandic19gcpr}
Petra Bevandi{\'c}, Ivan Kre{\v{s}}o, Marin Or{\v{s}}i{\'c}, and Sini{\v{s}}a {\v{S}}egvi{\'c}.
\newblock Simultaneous semantic segmentation and outlier detection in presence of domain shift.
\newblock In \emph{41st DAGM German Conference, DAGM GCPR}. Springer, 2019.

\bibitem[Bevandi{\'c} et~al.(2022)Bevandi{\'c}, Kre{\v{s}}o, Or{\v{s}}i{\'c}, and {\v{S}}egvi{\'c}]{bevandic22ivc}
Petra Bevandi{\'c}, Ivan Kre{\v{s}}o, Marin Or{\v{s}}i{\'c}, and Sini{\v{s}}a {\v{S}}egvi{\'c}.
\newblock Dense open-set recognition based on training with noisy negative images.
\newblock \emph{Image and Vision Computing}, 2022.

\bibitem[Biase et~al.(2021)Biase, Blum, Siegwart, and Cadena]{biase2021cvpr}
Giancarlo~Di Biase, Hermann Blum, Roland Siegwart, and C{\'{e}}sar Cadena.
\newblock Pixel-wise anomaly detection in complex driving scenes.
\newblock In \emph{{IEEE} Conference on Computer Vision and Pattern Recognition, {CVPR} 2021, virtual, June 19-25, 2021}, pages 16918--16927. Computer Vision Foundation / {IEEE}, 2021.
\newblock \doi{10.1109/CVPR46437.2021.01664}.

\bibitem[Bitterwolf et~al.(2023)Bitterwolf, M{\"u}ller, and Hein]{bitterwolf2023or}
Julian Bitterwolf, Maximilian M{\"u}ller, and Matthias Hein.
\newblock In or out? fixing imagenet out-of-distribution detection evaluation.
\newblock \emph{arXiv preprint arXiv:2306.00826}, 2023.

\bibitem[Blum et~al.(2021)Blum, Sarlin, Nieto, Siegwart, and Cadena]{blum2021fishyscapes}
Hermann Blum, Paul-Edouard Sarlin, Juan Nieto, Roland Siegwart, and Cesar Cadena.
\newblock The fishyscapes benchmark: Measuring blind spots in semantic segmentation.
\newblock \emph{International Journal of Computer Vision}, 2021.

\bibitem[Carion et~al.(2020)Carion, Massa, Synnaeve, Usunier, Kirillov, and Zagoruyko]{carion2020eccv}
Nicolas Carion, Francisco Massa, Gabriel Synnaeve, Nicolas Usunier, Alexander Kirillov, and Sergey Zagoruyko.
\newblock End-to-end object detection with transformers.
\newblock In Andrea Vedaldi, Horst Bischof, Thomas Brox, and Jan{-}Michael Frahm, editors, \emph{Computer Vision - {ECCV} 2020 - 16th European Conference, Glasgow, UK, August 23-28, 2020, Proceedings, Part {I}}, volume 12346 of \emph{Lecture Notes in Computer Science}, pages 213--229. Springer, 2020.

\bibitem[Chan et~al.(2021{\natexlab{a}})Chan, Lis, Uhlemeyer, Blum, Honari, Siegwart, Fua, Salzmann, and Rottmann]{chan2021segmentmeifyoucan}
Robin Chan, Krzysztof Lis, Svenja Uhlemeyer, Hermann Blum, Sina Honari, Roland Siegwart, Pascal Fua, Mathieu Salzmann, and Matthias Rottmann.
\newblock Segmentmeifyoucan: A benchmark for anomaly segmentation.
\newblock \emph{arXiv preprint arXiv:2104.14812}, 2021{\natexlab{a}}.

\bibitem[Chan et~al.(2021{\natexlab{b}})Chan, Rottmann, and Gottschalk]{chan2021entropy}
Robin Chan, Matthias Rottmann, and Hanno Gottschalk.
\newblock Entropy maximization and meta classification for out-of-distribution detection in semantic segmentation.
\newblock In \emph{Proceedings of the ieee/cvf international conference on computer vision}, pages 5128--5137, 2021{\natexlab{b}}.

\bibitem[Chen et~al.(2022)Chen, Peng, Wang, and Tian]{chen2022pami}
Guangyao Chen, Peixi Peng, Xiangqian Wang, and Yonghong Tian.
\newblock Adversarial reciprocal points learning for open set recognition.
\newblock \emph{{IEEE} Trans. Pattern Anal. Mach. Intell.}, 44\penalty0 (11):\penalty0 8065--8081, 2022.

\bibitem[Chen et~al.(2018)Chen, Papandreou, Kokkinos, Murphy, and Yuille]{chen18tpami}
Liang{-}Chieh Chen, George Papandreou, Iasonas Kokkinos, Kevin Murphy, and Alan~L. Yuille.
\newblock Deeplab: Semantic image segmentation with deep convolutional nets, atrous convolution, and fully connected crfs.
\newblock \emph{{IEEE} Trans. Pattern Anal. Mach. Intell.}, 40\penalty0 (4):\penalty0 834--848, 2018.

\bibitem[Chen et~al.(2023)Chen, Lin, Xu, and Vela]{chen2023iccv}
Yiye Chen, Yunzhi Lin, Ruinian Xu, and Patricio~A. Vela.
\newblock Wdiscood: Out-of-distribution detection via whitened linear discriminant analysis.
\newblock In \emph{{IEEE/CVF} International Conference on Computer Vision, {ICCV} 2023, Paris, France, October 1-6, 2023}, pages 5275--5284. {IEEE}, 2023.

\bibitem[Cheng et~al.(2022)Cheng, Misra, Schwing, Kirillov, and Girdhar]{cheng2022masked}
Bowen Cheng, Ishan Misra, Alexander~G Schwing, Alexander Kirillov, and Rohit Girdhar.
\newblock Masked-attention mask transformer for universal image segmentation.
\newblock In \emph{Proceedings of the IEEE/CVF conference on computer vision and pattern recognition}, pages 1290--1299, 2022.

\bibitem[Cimpoi et~al.(2014)Cimpoi, Maji, Kokkinos, Mohamed, and Vedaldi]{cimpoi2014describing}
Mircea Cimpoi, Subhransu Maji, Iasonas Kokkinos, Sammy Mohamed, and Andrea Vedaldi.
\newblock Describing textures in the wild.
\newblock In \emph{Proceedings of the IEEE conference on computer vision and pattern recognition}, pages 3606--3613, 2014.

\bibitem[Cordts et~al.(2016)Cordts, Omran, Ramos, Rehfeld, Enzweiler, Benenson, Franke, Roth, and Schiele]{cordts2016cityscapes}
Marius Cordts, Mohamed Omran, Sebastian Ramos, Timo Rehfeld, Markus Enzweiler, Rodrigo Benenson, Uwe Franke, Stefan Roth, and Bernt Schiele.
\newblock The cityscapes dataset for semantic urban scene understanding.
\newblock In \emph{Proceedings of the IEEE conference on computer vision and pattern recognition}, pages 3213--3223, 2016.

\bibitem[Deng et~al.(2009)Deng, Dong, Socher, Li, Li, and Fei{-}Fei]{deng09cvpr}
Jia Deng, Wei Dong, Richard Socher, Li{-}Jia Li, Kai Li, and Li~Fei{-}Fei.
\newblock Imagenet: {A} large-scale hierarchical image database.
\newblock In \emph{{IEEE} Computer Society Conference on Computer Vision and Pattern Recognition {CVPR}}, 2009.

\bibitem[Deng(2012)]{deng2012mnist}
Li~Deng.
\newblock The mnist database of handwritten digit images for machine learning research [best of the web].
\newblock \emph{IEEE signal processing magazine}, 29\penalty0 (6):\penalty0 141--142, 2012.

\bibitem[DeVries and Taylor(2018)]{devries2018arxiv}
Terrance DeVries and Graham~W. Taylor.
\newblock Learning confidence for out-of-distribution detection in neural networks.
\newblock \emph{CoRR}, abs/1802.04865, 2018.
\newblock URL \url{http://arxiv.org/abs/1802.04865}.

\bibitem[Dhamija et~al.(2018)Dhamija, G\"{u}nther, and Boult]{dhamija18neurips}
Akshay~Raj Dhamija, Manuel G\"{u}nther, and Terrance~E. Boult.
\newblock Reducing network agnostophobia.
\newblock In \emph{Proceedings of the 32nd International Conference on Neural Information Processing Systems}, NIPS'18, page 9175–9186, 2018.

\bibitem[Di~Biase et~al.(2021)Di~Biase, Blum, Siegwart, and Cadena]{di2021pixel}
Giancarlo Di~Biase, Hermann Blum, Roland Siegwart, and Cesar Cadena.
\newblock Pixel-wise anomaly detection in complex driving scenes.
\newblock In \emph{Proceedings of the IEEE/CVF conference on computer vision and pattern recognition}, pages 16918--16927, 2021.

\bibitem[Dinh et~al.(2017)Dinh, Sohl-Dickstein, and Bengio]{dinh17iclr}
Laurent Dinh, Jascha Sohl-Dickstein, and Samy Bengio.
\newblock Density estimation using real {NVP}.
\newblock In \emph{International Conference on Learning Representations}, 2017.

\bibitem[Djurisic et~al.(2022)Djurisic, Bozanic, Ashok, and Liu]{djurisic2022extremely}
Andrija Djurisic, Nebojsa Bozanic, Arjun Ashok, and Rosanne Liu.
\newblock Extremely simple activation shaping for out-of-distribution detection.
\newblock \emph{arXiv preprint arXiv:2209.09858}, 2022.

\bibitem[Dosovitskiy et~al.(2021)Dosovitskiy, Beyer, Kolesnikov, Weissenborn, Zhai, Unterthiner, Dehghani, Minderer, Heigold, Gelly, Uszkoreit, and Houlsby]{dosovitskiy21iclr}
Alexey Dosovitskiy, Lucas Beyer, Alexander Kolesnikov, Dirk Weissenborn, Xiaohua Zhai, Thomas Unterthiner, Mostafa Dehghani, Matthias Minderer, Georg Heigold, Sylvain Gelly, Jakob Uszkoreit, and Neil Houlsby.
\newblock An image is worth 16x16 words: Transformers for image recognition at scale.
\newblock In \emph{International Conference on Learning Representations, {ICLR}}, 2021.

\bibitem[Du et~al.(2022)Du, Wang, Cai, and Li]{du2022iclr}
Xuefeng Du, Zhaoning Wang, Mu~Cai, and Yixuan Li.
\newblock {VOS:} learning what you don't know by virtual outlier synthesis.
\newblock In \emph{The Tenth International Conference on Learning Representations, {ICLR}}, 2022.

\bibitem[Galesso et~al.(2022)Galesso, Argus, and Brox]{galesso23iccv}
Silvio Galesso, Max Argus, and Thomas Brox.
\newblock Far away in the deep space: Nearest-neighbor-based dense out-of-distribution detection.
\newblock \emph{CoRR}, abs/2211.06660, 2022.
\newblock \doi{10.48550/ARXIV.2211.06660}.
\newblock URL \url{https://doi.org/10.48550/arXiv.2211.06660}.

\bibitem[Grci{\'c} et~al.(2021{\natexlab{a}})Grci{\'c}, Bevandi{\'c}, and {\v{S}}egvi{\'c}]{grcic21arxiv}
Matej Grci{\'c}, Petra Bevandi{\'c}, and Sini{\v{s}}a {\v{S}}egvi{\'c}.
\newblock Dense anomaly detection by robust learning on synthetic negative data.
\newblock \emph{arXiv preprint arXiv:2112.12833}, 2021{\natexlab{a}}.

\bibitem[Grci{\'c} et~al.(2021{\natexlab{b}})Grci{\'c}, Bevandi{\'c}, and {\v{S}}egvi{\'c}]{grcic21visapp}
Matej Grci{\'c}, Petra Bevandi{\'c}, and Sini{\v{s}}a {\v{S}}egvi{\'c}.
\newblock Dense open-set recognition with synthetic outliers generated by real nvp.
\newblock In \emph{16th International Joint Conference on Computer Vision, Imaging and Computer Graphics Theory and Applications, VISIGRAPP}, 2021{\natexlab{b}}.

\bibitem[Grci{\'c} et~al.(2021{\natexlab{c}})Grci{\'c}, Grubi{\v{s}}i{\'c}, and {\v{S}}egvi{\'c}]{grcic21neurips}
Matej Grci{\'c}, Ivan Grubi{\v{s}}i{\'c}, and Sini{\v{s}}a {\v{S}}egvi{\'c}.
\newblock Densely connected normalizing flows.
\newblock \emph{Advances in Neural Information Processing Systems}, 2021{\natexlab{c}}.

\bibitem[Grcic et~al.(2022)Grcic, Bevandic, and Segvic]{grcic22eccv}
Matej Grcic, Petra Bevandic, and Sinisa Segvic.
\newblock Densehybrid: Hybrid anomaly detection for dense open-set recognition.
\newblock In \emph{European Conference on Computer Vision, ECCV 2022}. Springer, 2022.

\bibitem[Grci{\'c} et~al.(2023)Grci{\'c}, {\v{S}}ari{\'c}, and {\v{S}}egvi{\'c}]{grcic2023advantages}
Matej Grci{\'c}, Josip {\v{S}}ari{\'c}, and Sini{\v{s}}a {\v{S}}egvi{\'c}.
\newblock On advantages of mask-level recognition for outlier-aware segmentation.
\newblock In \emph{Proceedings of the IEEE/CVF Conference on Computer Vision and Pattern Recognition}, pages 2936--2946, 2023.

\bibitem[Grcic et~al.(2024)Grcic, Bevandic, Kalafatic, and Segvic]{grcic24sensor}
Matej Grcic, Petra Bevandic, Zoran Kalafatic, and Sinisa Segvic.
\newblock Dense out-of-distribution detection by robust learning on synthetic negative data.
\newblock \emph{Sensors}, 24\penalty0 (4):\penalty0 1248, 2024.

\bibitem[Grcić and Šegvić(2024{\natexlab{a}})]{grcic23arxiv}
Matej Grcić and Siniša Šegvić.
\newblock Hybrid open-set segmentation with synthetic negative data.
\newblock \emph{IEEE Transactions on Pattern Analysis and Machine Intelligence}, 2024{\natexlab{a}}.

\bibitem[Grcić and Šegvić(2024{\natexlab{b}})]{grcic24tpami}
Matej Grcić and Siniša Šegvić.
\newblock Hybrid open-set segmentation with synthetic negative data.
\newblock \emph{IEEE Transactions on Pattern Analysis and Machine Intelligence}, 2024{\natexlab{b}}.

\bibitem[Guo et~al.(2017)Guo, Pleiss, Sun, and Weinberger]{guo2017calibration}
Chuan Guo, Geoff Pleiss, Yu~Sun, and Kilian~Q Weinberger.
\newblock On calibration of modern neural networks.
\newblock In \emph{International conference on machine learning}, pages 1321--1330. PMLR, 2017.

\bibitem[He et~al.(2016)He, Zhang, Ren, and Sun]{he2016deep}
Kaiming He, Xiangyu Zhang, Shaoqing Ren, and Jian Sun.
\newblock Deep residual learning for image recognition.
\newblock In \emph{Proceedings of the IEEE conference on computer vision and pattern recognition}, pages 770--778, 2016.

\bibitem[Hendrycks and Gimpel(2016)]{hendrycks2016baseline}
Dan Hendrycks and Kevin Gimpel.
\newblock A baseline for detecting misclassified and out-of-distribution examples in neural networks.
\newblock \emph{arXiv preprint arXiv:1610.02136}, 2016.

\bibitem[Hendrycks and Gimpel(2017)]{hendrycks17iclr}
Dan Hendrycks and Kevin Gimpel.
\newblock A baseline for detecting misclassified and out-of-distribution examples in neural networks.
\newblock In \emph{5th International Conference on Learning Representations, {ICLR} 2017, Toulon, France, April 24-26, 2017, Conference Track Proceedings}. OpenReview.net, 2017.

\bibitem[Hendrycks et~al.(2018)Hendrycks, Mazeika, and Dietterich]{hendrycks2018deep}
Dan Hendrycks, Mantas Mazeika, and Thomas Dietterich.
\newblock Deep anomaly detection with outlier exposure.
\newblock \emph{arXiv preprint arXiv:1812.04606}, 2018.

\bibitem[Hendrycks et~al.(2019{\natexlab{a}})Hendrycks, Basart, Mazeika, Zou, Kwon, Mostajabi, Steinhardt, and Song]{hendrycks2019scaling}
Dan Hendrycks, Steven Basart, Mantas Mazeika, Andy Zou, Joe Kwon, Mohammadreza Mostajabi, Jacob Steinhardt, and Dawn Song.
\newblock Scaling out-of-distribution detection for real-world settings.
\newblock \emph{arXiv preprint arXiv:1911.11132}, 2019{\natexlab{a}}.

\bibitem[Hendrycks et~al.(2019{\natexlab{b}})Hendrycks, Mazeika, Kadavath, and Song]{hendrycks2019using}
Dan Hendrycks, Mantas Mazeika, Saurav Kadavath, and Dawn Song.
\newblock Using self-supervised learning can improve model robustness and uncertainty.
\newblock \emph{Advances in neural information processing systems}, 32, 2019{\natexlab{b}}.

\bibitem[Hendrycks et~al.(2022{\natexlab{a}})Hendrycks, Basart, Mazeika, Zou, Kwon, Mostajabi, Steinhardt, and Song]{hendrycks19icml}
Dan Hendrycks, Steven Basart, Mantas Mazeika, Andy Zou, Joe Kwon, Mohammadreza Mostajabi, Jacob Steinhardt, and Dawn Song.
\newblock Scaling out-of-distribution detection for real-world settings.
\newblock \emph{ICML}, 2022{\natexlab{a}}.

\bibitem[Hendrycks et~al.(2022{\natexlab{b}})Hendrycks, Basart, Mazeika, Zou, Kwon, Mostajabi, Steinhardt, and Song]{hendrycks2022icml}
Dan Hendrycks, Steven Basart, Mantas Mazeika, Andy Zou, Joseph Kwon, Mohammadreza Mostajabi, Jacob Steinhardt, and Dawn Song.
\newblock Scaling out-of-distribution detection for real-world settings.
\newblock In Kamalika Chaudhuri, Stefanie Jegelka, Le~Song, Csaba Szepesv{\'{a}}ri, Gang Niu, and Sivan Sabato, editors, \emph{International Conference on Machine Learning, {ICML} 2022, 17-23 July 2022, Baltimore, Maryland, {USA}}, volume 162 of \emph{Proceedings of Machine Learning Research}, pages 8759--8773. {PMLR}, 2022{\natexlab{b}}.

\bibitem[Hinton et~al.(2015)Hinton, Vinyals, and Dean]{hinton2015arxiv}
Geoffrey~E. Hinton, Oriol Vinyals, and Jeffrey Dean.
\newblock Distilling the knowledge in a neural network.
\newblock \emph{CoRR}, abs/1503.02531, 2015.

\bibitem[Hsu et~al.(2020)Hsu, Shen, Jin, and Kira]{hsu2020cvpr}
Yen{-}Chang Hsu, Yilin Shen, Hongxia Jin, and Zsolt Kira.
\newblock Generalized {ODIN:} detecting out-of-distribution image without learning from out-of-distribution data.
\newblock In \emph{2020 {IEEE/CVF} Conference on Computer Vision and Pattern Recognition, {CVPR} 2020, Seattle, WA, USA, June 13-19, 2020}, pages 10948--10957. Computer Vision Foundation / {IEEE}, 2020.

\bibitem[Huang and Li(2021)]{huang2021cvpr}
Rui Huang and Yixuan Li.
\newblock {MOS:} towards scaling out-of-distribution detection for large semantic space.
\newblock In \emph{{IEEE} Conference on Computer Vision and Pattern Recognition, {CVPR} 2021, virtual, June 19-25, 2021}, pages 8710--8719. Computer Vision Foundation / {IEEE}, 2021.

\bibitem[Huang et~al.(2021)Huang, Geng, and Li]{huang2021neurips}
Rui Huang, Andrew Geng, and Yixuan Li.
\newblock On the importance of gradients for detecting distributional shifts in the wild.
\newblock In Marc'Aurelio Ranzato, Alina Beygelzimer, Yann~N. Dauphin, Percy Liang, and Jennifer~Wortman Vaughan, editors, \emph{Advances in Neural Information Processing Systems 34: Annual Conference on Neural Information Processing Systems 2021, NeurIPS 2021, December 6-14, 2021, virtual}, pages 677--689, 2021.

\bibitem[Jung et~al.(2021)Jung, Lee, Gwak, Choi, and Choo]{jung2021standardized}
Sanghun Jung, Jungsoo Lee, Daehoon Gwak, Sungha Choi, and Jaegul Choo.
\newblock Standardized max logits: A simple yet effective approach for identifying unexpected road obstacles in urban-scene segmentation.
\newblock In \emph{Proceedings of the IEEE/CVF International Conference on Computer Vision}, pages 15425--15434, 2021.

\bibitem[Kingma and Dhariwal(2018)]{kingma2018neurips}
Diederik~P. Kingma and Prafulla Dhariwal.
\newblock Glow: Generative flow with invertible 1x1 convolutions.
\newblock In \emph{Advances in Neural Information Processing Systems 31: Annual Conference on Neural Information Processing Systems 2018, NeurIPS 2018}, 2018.

\bibitem[Kong and Ramanan(2021)]{kong2021iccv}
Shu Kong and Deva Ramanan.
\newblock Opengan: Open-set recognition via open data generation.
\newblock In \emph{2021 {IEEE/CVF} International Conference on Computer Vision, {ICCV} 2021, Montreal, QC, Canada, October 10-17, 2021}, pages 793--802. {IEEE}, 2021.

\bibitem[Krizhevsky(2009)]{krizhevsky2009learning}
Alex Krizhevsky.
\newblock Learning multiple layers of features from tiny images.
\newblock Technical report, University of Toronto, 2009.

\bibitem[Krizhevsky et~al.(2009)Krizhevsky, Nair, and Hinton]{krizhevsky2009cifar}
Alex Krizhevsky, Vinod Nair, and Geoffrey Hinton.
\newblock Cifar-10 and cifar-100 datasets.
\newblock \emph{URl: https://www. cs. toronto. edu/kriz/cifar. html}, 6\penalty0 (1):\penalty0 1, 2009.

\bibitem[Kumar et~al.(2023)Kumar, Segvic, Eslami, and Gumhold]{kumar23cvpr}
Nishant Kumar, Sinisa Segvic, Abouzar Eslami, and Stefan Gumhold.
\newblock Normalizing flow based feature synthesis for outlier-aware object detection.
\newblock In \emph{{IEEE/CVF} Conference on Computer Vision and Pattern Recognition, {CVPR} 2023, Vancouver, BC, Canada, June 17-24, 2023}, pages 5156--5165. {IEEE}, 2023.

\bibitem[Kuncheva and Whitaker(2003)]{kuncheva03ml}
L.~I. Kuncheva and C.~J. Whitaker.
\newblock Measures of diversity in classifier ensembles and their relationship with the ensemble accuracy.
\newblock \emph{Machine Learning}, 2003.

\bibitem[Lakshminarayanan et~al.(2017)Lakshminarayanan, Pritzel, and Blundell]{lakshminarayanan2017simple}
Balaji Lakshminarayanan, Alexander Pritzel, and Charles Blundell.
\newblock Simple and scalable predictive uncertainty estimation using deep ensembles.
\newblock \emph{Advances in neural information processing systems}, 30, 2017.

\bibitem[Le and Yang(2015)]{le2015tiny}
Ya~Le and Xuan Yang.
\newblock Tiny imagenet visual recognition challenge.
\newblock \emph{CS 231N}, 7\penalty0 (7):\penalty0 3, 2015.

\bibitem[Lee et~al.(2018{\natexlab{a}})Lee, Lee, Lee, and Shin]{lee18iclr}
Kimin Lee, Honglak Lee, Kibok Lee, and Jinwoo Shin.
\newblock Training confidence-calibrated classifiers for detecting out-of-distribution samples.
\newblock In \emph{International Conference on Learning Representations}, 2018{\natexlab{a}}.

\bibitem[Lee et~al.(2018{\natexlab{b}})Lee, Lee, Lee, and Shin]{Lee2018neurips}
Kimin Lee, Kibok Lee, Honglak Lee, and Jinwoo Shin.
\newblock A simple unified framework for detecting out-of-distribution samples and adversarial attacks.
\newblock In Samy Bengio, Hanna~M. Wallach, Hugo Larochelle, Kristen Grauman, Nicol{\`{o}} Cesa{-}Bianchi, and Roman Garnett, editors, \emph{Advances in Neural Information Processing Systems 31: Annual Conference on Neural Information Processing Systems 2018, NeurIPS 2018, December 3-8, 2018, Montr{\'{e}}al, Canada}, pages 7167--7177, 2018{\natexlab{b}}.

\bibitem[Li et~al.(2023)Li, Zhang, xu, Liu, Zhang, Ni, and Shum]{li23cvpr}
Feng Li, Hao Zhang, Huaizhe xu, Shilong Liu, Lei Zhang, Lionel~M. Ni, and Heung-Yeung Shum.
\newblock Mask dino: Towards a unified transformer-based framework for object detection and segmentation.
\newblock In \emph{CVPR}, 2023.

\bibitem[Liang et~al.(2022)Liang, Wang, Miao, and Yang]{liang22neurips}
Chen Liang, Wenguan Wang, Jiaxu Miao, and Yi~Yang.
\newblock Gmmseg: Gaussian mixture based generative semantic segmentation models.
\newblock In Sanmi Koyejo, S.~Mohamed, A.~Agarwal, Danielle Belgrave, K.~Cho, and A.~Oh, editors, \emph{Advances in Neural Information Processing Systems 35: Annual Conference on Neural Information Processing Systems 2022, NeurIPS 2022, New Orleans, LA, USA, November 28 - December 9, 2022}, 2022.

\bibitem[Liang et~al.(2017)Liang, Li, and Srikant]{liang2017enhancing}
Shiyu Liang, Yixuan Li, and Rayadurgam Srikant.
\newblock Enhancing the reliability of out-of-distribution image detection in neural networks.
\newblock \emph{arXiv preprint arXiv:1706.02690}, 2017.

\bibitem[Lin et~al.(2014{\natexlab{a}})Lin, Maire, Belongie, Hays, Perona, Ramanan, Doll{\'a}r, and Zitnick]{lin2014microsoft}
Tsung-Yi Lin, Michael Maire, Serge Belongie, James Hays, Pietro Perona, Deva Ramanan, Piotr Doll{\'a}r, and C~Lawrence Zitnick.
\newblock Microsoft coco: Common objects in context.
\newblock In \emph{Computer Vision--ECCV 2014: 13th European Conference, Zurich, Switzerland, September 6-12, 2014, Proceedings, Part V 13}, pages 740--755. Springer, 2014{\natexlab{a}}.

\bibitem[Lin et~al.(2014{\natexlab{b}})Lin, Maire, Belongie, Hays, Perona, Ramanan, Doll{\'{a}}r, and Zitnick]{lin14eccv}
Tsung{-}Yi Lin, Michael Maire, Serge~J. Belongie, James Hays, Pietro Perona, Deva Ramanan, Piotr Doll{\'{a}}r, and C.~Lawrence Zitnick.
\newblock Microsoft {COCO:} common objects in context.
\newblock In \emph{13th European Conference of Computer Vision {ECCV}}. Springer, 2014{\natexlab{b}}.

\bibitem[Lis et~al.(2019)Lis, Nakka, Fua, and Salzmann]{lis2019detecting}
Krzysztof Lis, Krishna Nakka, Pascal Fua, and Mathieu Salzmann.
\newblock Detecting the unexpected via image resynthesis.
\newblock In \emph{Proceedings of the IEEE/CVF International Conference on Computer Vision}, pages 2152--2161, 2019.

\bibitem[Liu et~al.(2020)Liu, Wang, Owens, and Li]{liu2020neurips}
Weitang Liu, Xiaoyun Wang, John~D. Owens, and Yixuan Li.
\newblock Energy-based out-of-distribution detection.
\newblock In Hugo Larochelle, Marc'Aurelio Ranzato, Raia Hadsell, Maria{-}Florina Balcan, and Hsuan{-}Tien Lin, editors, \emph{Advances in Neural Information Processing Systems 33: Annual Conference on Neural Information Processing Systems 2020, NeurIPS 2020, December 6-12, 2020, virtual}, 2020.

\bibitem[Liu et~al.(2023)Liu, Ding, Tian, Pang, Belagiannis, Reid, and Carneiro]{liu2023iccv}
Yuyuan Liu, Choubo Ding, Yu~Tian, Guansong Pang, Vasileios Belagiannis, Ian~D. Reid, and Gustavo Carneiro.
\newblock Residual pattern learning for pixel-wise out-of-distribution detection in semantic segmentation.
\newblock In \emph{{IEEE/CVF} International Conference on Computer Vision, {ICCV} 2023, Paris, France, October 1-6, 2023}, pages 1151--1161. {IEEE}, 2023.

\bibitem[Liu et~al.(2021)Liu, Lin, Cao, Hu, Wei, Zhang, Lin, and Guo]{liu2021swin}
Ze~Liu, Yutong Lin, Yue Cao, Han Hu, Yixuan Wei, Zheng Zhang, Stephen Lin, and Baining Guo.
\newblock Swin transformer: Hierarchical vision transformer using shifted windows.
\newblock In \emph{Proceedings of the IEEE/CVF international conference on computer vision}, pages 10012--10022, 2021.

\bibitem[Liu et~al.(2022)Liu, Mao, Wu, Feichtenhofer, Darrell, and Xie]{liu2022cvpr}
Zhuang Liu, Hanzi Mao, Chao{-}Yuan Wu, Christoph Feichtenhofer, Trevor Darrell, and Saining Xie.
\newblock A convnet for the 2020s.
\newblock In \emph{{IEEE/CVF} Conference on Computer Vision and Pattern Recognition, {CVPR} 2022, New Orleans, LA, USA, June 18-24, 2022}, 2022.

\bibitem[Lucas et~al.(2019)Lucas, Shmelkov, Alahari, Schmid, and Verbeek]{lucas19neurips}
Thomas Lucas, Konstantin Shmelkov, Karteek Alahari, Cordelia Schmid, and Jakob Verbeek.
\newblock Adaptive density estimation for generative models.
\newblock In Hanna~M. Wallach, Hugo Larochelle, Alina Beygelzimer, Florence d'Alch{\'{e}}{-}Buc, Emily~B. Fox, and Roman Garnett, editors, \emph{Advances in Neural Information Processing Systems 32: Annual Conference on Neural Information Processing Systems 2019, NeurIPS 2019, December 8-14, 2019, Vancouver, BC, Canada}, pages 11993--12003, 2019.

\bibitem[Malinin and Gales(2018)]{malinin2018predictive}
Andrey Malinin and Mark Gales.
\newblock Predictive uncertainty estimation via prior networks.
\newblock \emph{Advances in neural information processing systems}, 31, 2018.

\bibitem[Ming et~al.(2023)Ming, Sun, Dia, and Li]{ming2023iclr}
Yifei Ming, Yiyou Sun, Ousmane Dia, and Yixuan Li.
\newblock How to exploit hyperspherical embeddings for out-of-distribution detection?
\newblock In \emph{The Eleventh International Conference on Learning Representations, {ICLR} 2023, Kigali, Rwanda, May 1-5, 2023}. OpenReview.net, 2023.

\bibitem[Mukhoti and Gal(2018)]{mukhoti2018evaluating}
Jishnu Mukhoti and Yarin Gal.
\newblock Evaluating bayesian deep learning methods for semantic segmentation.
\newblock \emph{arXiv preprint arXiv:1811.12709}, 2018.

\bibitem[Nayal et~al.(2023)Nayal, Yavuz, Henriques, and G{\"u}ney]{nayal2023rba}
Nazir Nayal, Misra Yavuz, Joao~F Henriques, and Fatma G{\"u}ney.
\newblock Rba: Segmenting unknown regions rejected by all.
\newblock In \emph{Proceedings of the IEEE/CVF International Conference on Computer Vision}, pages 711--722, 2023.

\bibitem[Neal et~al.(2018)Neal, Olson, Fern, Wong, and Li]{neal18eccv}
Lawrence Neal, Matthew Olson, Xiaoli Fern, Weng-Keen Wong, and Fuxin Li.
\newblock Open set learning with counterfactual images.
\newblock In \emph{Proceedings of the European Conference on Computer Vision (ECCV)}, 2018.

\bibitem[Neuhold et~al.(2017)Neuhold, Ollmann, Rota~Bulo, and Kontschieder]{neuhold2017mapillary}
Gerhard Neuhold, Tobias Ollmann, Samuel Rota~Bulo, and Peter Kontschieder.
\newblock The mapillary vistas dataset for semantic understanding of street scenes.
\newblock In \emph{Proceedings of the IEEE international conference on computer vision}, pages 4990--4999, 2017.

\bibitem[Noh et~al.(2023)Noh, Jeong, and Lee]{noh2023iccv}
SoonCheol Noh, DongEon Jeong, and Jee{-}Hyong Lee.
\newblock Simple and effective out-of-distribution detection via cosine-based softmax loss.
\newblock In \emph{{IEEE/CVF} International Conference on Computer Vision, {ICCV} 2023, Paris, France, October 1-6, 2023}, pages 16514--16523. {IEEE}, 2023.

\bibitem[Park et~al.(2023)Park, Jung, and Teoh]{park2023iccvb}
Jaewoo Park, Yoon~Gyo Jung, and Andrew Beng~Jin Teoh.
\newblock Nearest neighbor guidance for out-of-distribution detection.
\newblock In \emph{{IEEE/CVF} International Conference on Computer Vision, {ICCV} 2023, Paris, France, October 1-6, 2023}, pages 1686--1695. {IEEE}, 2023.

\bibitem[Rai et~al.(2023)Rai, Cermelli, Fontanel, Masone, and Caputo]{rai2023unmasking}
Shyam~Nandan Rai, Fabio Cermelli, Dario Fontanel, Carlo Masone, and Barbara Caputo.
\newblock Unmasking anomalies in road-scene segmentation.
\newblock In \emph{Proceedings of the IEEE/CVF International Conference on Computer Vision}, pages 4037--4046, 2023.

\bibitem[Ren et~al.(2021)Ren, Fort, Liu, Roy, Padhy, and Lakshminarayanan]{ren2021arxiv}
Jie Ren, Stanislav Fort, Jeremiah~Z. Liu, Abhijit~Guha Roy, Shreyas Padhy, and Balaji Lakshminarayanan.
\newblock A simple fix to mahalanobis distance for improving near-ood detection.
\newblock \emph{CoRR}, abs/2106.09022, 2021.

\bibitem[Ruff et~al.(2021)Ruff, Kauffmann, Vandermeulen, Montavon, Samek, Kloft, Dietterich, and M{\"{u}}ller]{ruff21pieee}
Lukas Ruff, Jacob~R. Kauffmann, Robert~A. Vandermeulen, Gr{\'{e}}goire Montavon, Wojciech Samek, Marius Kloft, Thomas~G. Dietterich, and Klaus{-}Robert M{\"{u}}ller.
\newblock A unifying review of deep and shallow anomaly detection.
\newblock \emph{Proc. {IEEE}}, 109\penalty0 (5):\penalty0 756--795, 2021.

\bibitem[Sastry and Oore(2020)]{sastry2020icml}
Chandramouli~Shama Sastry and Sageev Oore.
\newblock Detecting out-of-distribution examples with gram matrices.
\newblock In \emph{Proceedings of the 37th International Conference on Machine Learning, {ICML} 2020, 13-18 July 2020, Virtual Event}, volume 119 of \emph{Proceedings of Machine Learning Research}, pages 8491--8501. {PMLR}, 2020.

\bibitem[Song et~al.(2022)Song, Sebe, and Wang]{song2022neurips}
Yue Song, Nicu Sebe, and Wei Wang.
\newblock Rankfeat: Rank-1 feature removal for out-of-distribution detection.
\newblock In Sanmi Koyejo, S.~Mohamed, A.~Agarwal, Danielle Belgrave, K.~Cho, and A.~Oh, editors, \emph{Advances in Neural Information Processing Systems 35: Annual Conference on Neural Information Processing Systems 2022, NeurIPS 2022, New Orleans, LA, USA, November 28 - December 9, 2022}, 2022.

\bibitem[Sun and Li(2022)]{sun2022eccv}
Yiyou Sun and Yixuan Li.
\newblock {DICE:} leveraging sparsification for out-of-distribution detection.
\newblock In Shai Avidan, Gabriel~J. Brostow, Moustapha Ciss{\'{e}}, Giovanni~Maria Farinella, and Tal Hassner, editors, \emph{Computer Vision - {ECCV} 2022: 17th European Conference, Tel Aviv, Israel, October 23-27, 2022, Proceedings, Part {XXIV}}, volume 13684 of \emph{Lecture Notes in Computer Science}, pages 691--708. Springer, 2022.

\bibitem[Sun et~al.(2021)Sun, Guo, and Li]{sun2021neurips}
Yiyou Sun, Chuan Guo, and Yixuan Li.
\newblock React: Out-of-distribution detection with rectified activations.
\newblock In Marc'Aurelio Ranzato, Alina Beygelzimer, Yann~N. Dauphin, Percy Liang, and Jennifer~Wortman Vaughan, editors, \emph{Advances in Neural Information Processing Systems 34: Annual Conference on Neural Information Processing Systems 2021, NeurIPS 2021, December 6-14, 2021, virtual}, pages 144--157, 2021.

\bibitem[Sun et~al.(2022)Sun, Ming, Zhu, and Li]{sun2022icml}
Yiyou Sun, Yifei Ming, Xiaojin Zhu, and Yixuan Li.
\newblock Out-of-distribution detection with deep nearest neighbors.
\newblock In Kamalika Chaudhuri, Stefanie Jegelka, Le~Song, Csaba Szepesv{\'{a}}ri, Gang Niu, and Sivan Sabato, editors, \emph{International Conference on Machine Learning, {ICML} 2022, 17-23 July 2022, Baltimore, Maryland, {USA}}, volume 162 of \emph{Proceedings of Machine Learning Research}, pages 20827--20840. {PMLR}, 2022.

\bibitem[Tack et~al.(2020)Tack, Mo, Jeong, and Shin]{tack2020neurips}
Jihoon Tack, Sangwoo Mo, Jongheon Jeong, and Jinwoo Shin.
\newblock {CSI:} novelty detection via contrastive learning on distributionally shifted instances.
\newblock In \emph{Advances in Neural Information Processing Systems 33: Annual Conference on Neural Information Processing Systems 2020, NeurIPS 2020, December 6-12, 2020, virtual}, 2020.

\bibitem[Taghikhah et~al.(2024)Taghikhah, Kumar, Segvic, Eslami, and Gumhold]{taghikah24aaai}
Masoud Taghikhah, Nishant Kumar, Sinisa Segvic, Abouzar Eslami, and Stefan Gumhold.
\newblock Quantile-based maximum likelihood training for outlier detection.
\newblock In Michael~J. Wooldridge, Jennifer~G. Dy, and Sriraam Natarajan, editors, \emph{Thirty-Eighth {AAAI} Conference on Artificial Intelligence, {AAAI} 2024, Thirty-Sixth Conference on Innovative Applications of Artificial Intelligence, {IAAI} 2024, Fourteenth Symposium on Educational Advances in Artificial Intelligence, {EAAI} 2014, February 20-27, 2024, Vancouver, Canada}, pages 21610--21618. {AAAI} Press, 2024.

\bibitem[Tao et~al.(2023)Tao, Du, Zhu, and Li]{tao2023iclr}
Leitian Tao, Xuefeng Du, Jerry Zhu, and Yixuan Li.
\newblock Non-parametric outlier synthesis.
\newblock In \emph{The Eleventh International Conference on Learning Representations, {ICLR}}, 2023.

\bibitem[Tian et~al.(2022)Tian, Liu, Pang, Liu, Chen, and Carneiro]{tian2022pixel}
Yu~Tian, Yuyuan Liu, Guansong Pang, Fengbei Liu, Yuanhong Chen, and Gustavo Carneiro.
\newblock Pixel-wise energy-biased abstention learning for anomaly segmentation on complex urban driving scenes.
\newblock In \emph{European Conference on Computer Vision}, pages 246--263. Springer, 2022.

\bibitem[Van~Horn et~al.(2018)Van~Horn, Mac~Aodha, Song, Cui, Sun, Shepard, Adam, Perona, and Belongie]{van2018inaturalist}
Grant Van~Horn, Oisin Mac~Aodha, Yang Song, Yin Cui, Chen Sun, Alex Shepard, Hartwig Adam, Pietro Perona, and Serge Belongie.
\newblock The inaturalist species classification and detection dataset.
\newblock In \emph{Proceedings of the IEEE conference on computer vision and pattern recognition}, pages 8769--8778, 2018.

\bibitem[Vaze et~al.(2021)Vaze, Han, Vedaldi, and Zisserman]{vaze2021open}
Sagar Vaze, Kai Han, Andrea Vedaldi, and Andrew Zisserman.
\newblock Open-set recognition: A good closed-set classifier is all you need?
\newblock \emph{arXiv preprint arXiv:2110.06207}, 2021.

\bibitem[Vojir et~al.(2021)Vojir, {\v{S}}ipka, Aljundi, Chumerin, Reino, and Matas]{vojir2021road}
Tomas Vojir, Tom{\'a}{\v{s}} {\v{S}}ipka, Rahaf Aljundi, Nikolay Chumerin, Daniel~Olmeda Reino, and Jiri Matas.
\newblock Road anomaly detection by partial image reconstruction with segmentation coupling.
\newblock In \emph{Proceedings of the IEEE/CVF International Conference on Computer Vision}, pages 15651--15660, 2021.

\bibitem[Wang et~al.(2022)Wang, Li, Feng, and Zhang]{wang2022vim}
Haoqi Wang, Zhizhong Li, Litong Feng, and Wayne Zhang.
\newblock Vim: Out-of-distribution with virtual-logit matching.
\newblock In \emph{Proceedings of the IEEE/CVF conference on computer vision and pattern recognition}, pages 4921--4930, 2022.

\bibitem[Wei et~al.(2022)Wei, Xie, Cheng, Feng, An, and Li]{wei2022icml}
Hongxin Wei, Renchunzi Xie, Hao Cheng, Lei Feng, Bo~An, and Yixuan Li.
\newblock Mitigating neural network overconfidence with logit normalization.
\newblock In \emph{International Conference on Machine Learning, {ICML} 2022, 17-23 July 2022}, volume 162 of \emph{Proceedings of Machine Learning Research}, pages 23631--23644. {PMLR}, 2022.

\bibitem[Yang et~al.(2021)Yang, Wang, Feng, Yan, Zheng, Zhang, and Liu]{yang2021semantically}
Jingkang Yang, Haoqi Wang, Litong Feng, Xiaopeng Yan, Huabin Zheng, Wayne Zhang, and Ziwei Liu.
\newblock Semantically coherent out-of-distribution detection.
\newblock In \emph{Proceedings of the IEEE/CVF International Conference on Computer Vision}, pages 8301--8309, 2021.

\bibitem[Yu et~al.(2022)Yu, Wang, Qiao, Collins, Zhu, Adam, Yuille, and Chen]{yu22eccv}
Qihang Yu, Huiyu Wang, Siyuan Qiao, Maxwell Collins, Yukun Zhu, Hartwig Adam, Alan Yuille, and Liang-Chieh Chen.
\newblock {k-means Mask Transformer}.
\newblock In \emph{ECCV}, 2022.

\bibitem[Yu and Aizawa(2019)]{yu2019unsupervised}
Qing Yu and Kiyoharu Aizawa.
\newblock Unsupervised out-of-distribution detection by maximum classifier discrepancy.
\newblock In \emph{Proceedings of the IEEE/CVF international conference on computer vision}, pages 9518--9526, 2019.

\bibitem[Yuval(2011)]{yuval2011reading}
Netzer Yuval.
\newblock Reading digits in natural images with unsupervised feature learning.
\newblock In \emph{Proceedings of the NIPS Workshop on Deep Learning and Unsupervised Feature Learning}, 2011.

\bibitem[Zendel et~al.(2018)Zendel, Honauer, Murschitz, Steininger, and Dom{\'{\i}}nguez]{zendel18eccv}
Oliver Zendel, Katrin Honauer, Markus Murschitz, Daniel Steininger, and Gustavo~Fern{\'{a}}ndez Dom{\'{\i}}nguez.
\newblock Wilddash - creating hazard-aware benchmarks.
\newblock In Vittorio Ferrari, Martial Hebert, Cristian Sminchisescu, and Yair Weiss, editors, \emph{Computer Vision - {ECCV} 2018 - 15th European Conference, Munich, Germany, September 8-14, 2018, Proceedings, Part {VI}}, volume 11210 of \emph{Lecture Notes in Computer Science}, pages 407--421. Springer, 2018.
\newblock URL \url{https://doi.org/10.1007/978-3-030-01231-1\_25}.

\bibitem[Zhang et~al.(2020)Zhang, Li, Guo, and Guo]{zhang20eccv}
Hongjie Zhang, Ang Li, Jie Guo, and Yanwen Guo.
\newblock Hybrid models for open set recognition.
\newblock In Andrea Vedaldi, Horst Bischof, Thomas Brox, and Jan{-}Michael Frahm, editors, \emph{Computer Vision - {ECCV} 2020 - 16th European Conference, Glasgow, UK, August 23-28, 2020, Proceedings, Part {III}}, volume 12348 of \emph{Lecture Notes in Computer Science}, pages 102--117. Springer, 2020.

\bibitem[Zhang et~al.(2023{\natexlab{a}})Zhang, Inkawhich, Linderman, Chen, and Li]{zhang2023mixture}
Jingyang Zhang, Nathan Inkawhich, Randolph Linderman, Yiran Chen, and Hai Li.
\newblock Mixture outlier exposure: Towards out-of-distribution detection in fine-grained environments.
\newblock In \emph{Proceedings of the IEEE/CVF Winter Conference on Applications of Computer Vision}, pages 5531--5540, 2023{\natexlab{a}}.

\bibitem[Zhang et~al.(2023{\natexlab{b}})Zhang, Yang, Wang, Wang, Lin, Zhang, Sun, Du, Zhou, Zhang, et~al.]{zhang2023openood}
Jingyang Zhang, Jingkang Yang, Pengyun Wang, Haoqi Wang, Yueqian Lin, Haoran Zhang, Yiyou Sun, Xuefeng Du, Kaiyang Zhou, Wayne Zhang, et~al.
\newblock Openood v1. 5: Enhanced benchmark for out-of-distribution detection.
\newblock \emph{arXiv preprint arXiv:2306.09301}, 2023{\natexlab{b}}.

\bibitem[Zhang et~al.(2022)Zhang, Fu, Chen, Du, Li, Wang, Han, Zhang, et~al.]{zhang2022out}
Jinsong Zhang, Qiang Fu, Xu~Chen, Lun Du, Zelin Li, Gang Wang, Shi Han, Dongmei Zhang, et~al.
\newblock Out-of-distribution detection based on in-distribution data patterns memorization with modern hopfield energy.
\newblock In \emph{The Eleventh International Conference on Learning Representations}, 2022.

\bibitem[Zhang et~al.(2023{\natexlab{c}})Zhang, Fu, Chen, Du, Li, Wang, Liu, Han, and Zhang]{zhang2023iclr}
Jinsong Zhang, Qiang Fu, Xu~Chen, Lun Du, Zelin Li, Gang Wang, Xiaoguang Liu, Shi Han, and Dongmei Zhang.
\newblock Out-of-distribution detection based on in-distribution data patterns memorization with modern hopfield energy.
\newblock In \emph{The Eleventh International Conference on Learning Representations, {ICLR} 2023, Kigali, Rwanda, May 1-5, 2023}. OpenReview.net, 2023{\natexlab{c}}.

\bibitem[Zhao et~al.(2021)Zhao, Cao, and Lin]{zhao21arxiv}
Zhilin Zhao, Longbing Cao, and Kun{-}Yu Lin.
\newblock Revealing distributional vulnerability of explicit discriminators by implicit generators.
\newblock \emph{CoRR}, 2021.

\bibitem[Zhou et~al.(2017{\natexlab{a}})Zhou, Lapedriza, Khosla, Oliva, and Torralba]{zhou2017places}
Bolei Zhou, Agata Lapedriza, Aditya Khosla, Aude Oliva, and Antonio Torralba.
\newblock Places: A 10 million image database for scene recognition.
\newblock \emph{IEEE transactions on pattern analysis and machine intelligence}, 40\penalty0 (6):\penalty0 1452--1464, 2017{\natexlab{a}}.

\bibitem[Zhou et~al.(2017{\natexlab{b}})Zhou, Zhao, Puig, Fidler, Barriuso, and Torralba]{zhou2017scene}
Bolei Zhou, Hang Zhao, Xavier Puig, Sanja Fidler, Adela Barriuso, and Antonio Torralba.
\newblock Scene parsing through ade20k dataset.
\newblock In \emph{Proceedings of the IEEE conference on computer vision and pattern recognition}, 2017{\natexlab{b}}.

\bibitem[Zhou et~al.(2019)Zhou, Zhao, Puig, Xiao, Fidler, Barriuso, and Torralba]{zhou2019semantic}
Bolei Zhou, Hang Zhao, Xavier Puig, Tete Xiao, Sanja Fidler, Adela Barriuso, and Antonio Torralba.
\newblock Semantic understanding of scenes through the ade20k dataset.
\newblock \emph{International Journal of Computer Vision}, 127:\penalty0 302--321, 2019.

\end{thebibliography}

\newpage
\appendix
\section{Correlation analysis of UNO components}
Table~\ref{tab:corr} reports the Pearson correlation coefficient of per-pixel outlier scores $\textbf{s}_\text{Unc}$ and $\textbf{s}_\text{NO}$ on Fishyscapes \cite{blum2021fishyscapes} val and RoadAnomaly \cite{lis2019detecting}. 
We observe that the two UNO components are either mildly correlated or completely uncorrelated.
These findings indicate that the performance gains of UNO can be explained by the ensemble learning \cite{kuncheva03ml}.
\begin{table}[ht]
\small
\footnotesize
    \centering
    \begin{tabular}{c|ccc}
        Data & FS L\&F & FS Static & RoadAnomaly\\
         \hline
         Outliers & 0.27 & 0.01 & 0.13 \\
         Inliers & 0.01 & 0.15 & 0.01 \\
         All & 0.16 & 0.56 & 0.41 
    \end{tabular}
    \vskip 0.1in
    \caption{Pearson correlation coefficient of per-pixel scores $\textbf{s}_\text{Unc}$ and $\textbf{s}_\text{NO}$ on the three outlier segmentation validation datasets (FS L\&F, FS Static and RoadAnomaly).}
    \label{tab:corr}
\end{table}

\begin{figure}[ht]
    \centering
    \includegraphics[width=0.55\textwidth]{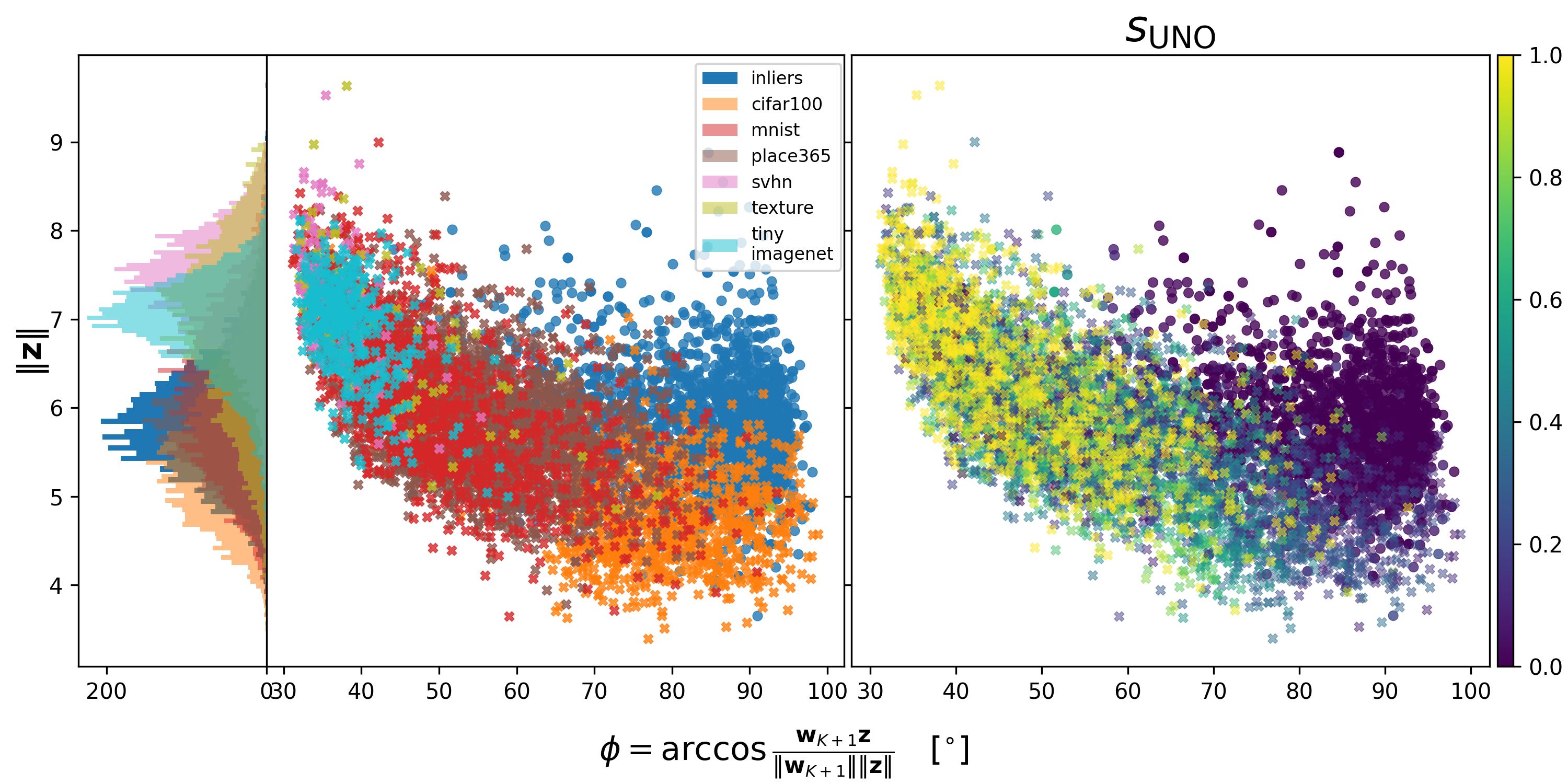}
    \caption{Visualization of the pre-logit space for  OpenOOD CIFAR-10.
    Outlier feature representations either yield an above-average norm with a small angle to $\mathbf{w}_{K+1}$ (e.g.~SVHN and Places365) or yield low norm representations (e.g.~CIFAR-100 that is similar to inliers).}
    \label{fig:norms_labeled}
\end{figure}

The benefits of ensembling can also be observed in the feature space.
Figure~\ref{fig:norms_labeled} indicates 
that features from different outlier datasets
have different $L_2$ norms.
Outliers that resemble the training negatives typically have a higher norm and small angle to $\mathbf{w}_{K+1}$, while outliers that are more similar to inliers have lower norm. In the former case, outliers are detected with $\textbf{s}_\text{NO}$ and with $\textbf{s}_\text{Unc}$ in the latter case.

\section{On orthogonality of class vectors}



\begin{figure}[ht]
    \centering
    \includegraphics[width=0.3\textwidth]{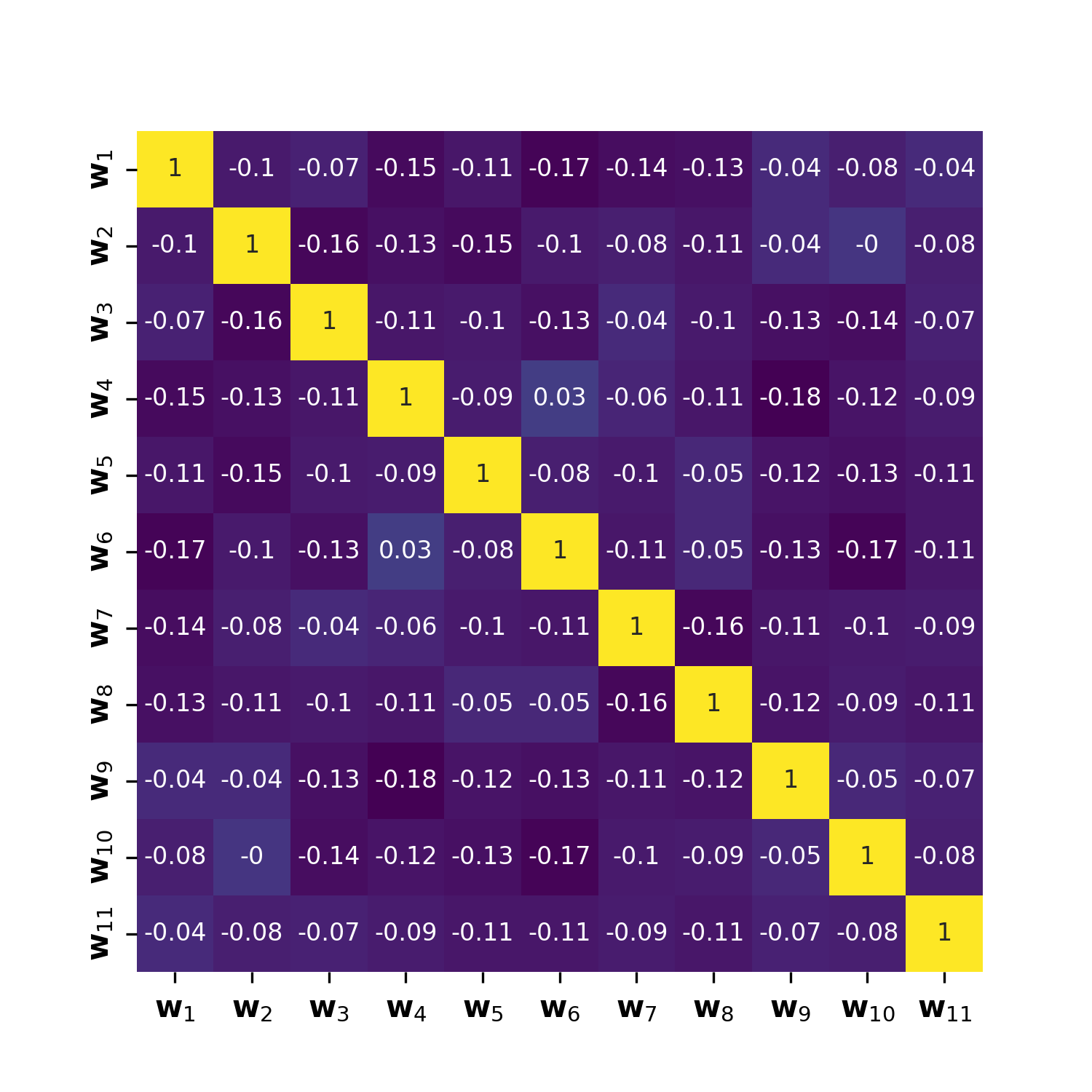}
    \caption{We show that all weight vectors are mutually orthogonal in the form of a heatmap where each element shows the cosine of the angle between two corresponding weight vectors. We use the K+1-way image-wide classifier with the ResNet-18 backbone trained on CIFAR-10 as the example.}
    \label{fig:orthogonal}
\end{figure}

We empirically observe that 
all class vectors $\mathbf{w}_i$ are mutually orthogonal to each other, $\mathbf{w}_i^T \mathbf{w}_j = 0, \,\, \forall i,j \in \{1, 2, ..., K+1\}$,
as
shown in Figure \ref{fig:orthogonal}. 
The cosine of the angle between any two different class vectors is approximately zero.
Such behaviour allows the geometrical interpretation of UNO, as shown in Figure 2 of the main manuscript. 

\section{Scene parsing with mask-wide recognition}
\label{m2f}
We extend the Mask2Former \cite{cheng2022masked} architecture with our UNO outlier detector to solve the task of anomaly segmentation. 
Thus, we describe Mask2Former architecture in detail to make the manuscript self-contained.
The Mask2Former architecture 
consists of three main parts: 
backbone, pixel decoder, and mask decoder.
The backbone extracts features 
at multiple scales from a given image 
$\mathbf{x} \in \mathbb{R}^{3\times H\times W}$. 
The pixel decoder produces 
high-resolution per-pixel features 
$\mathbf{E} \in \mathbb{R}^{E\times H\times W}$
that are fed into the mask decoder. 
The mask decoder  
formulates semantic segmentation
as a direct set prediction problem
by providing two outputs:
$N$ mask embeddings $\mathbf{q}$,
and $N$ mask-wide categorical 
distributions over K+1 classes 
$
  P (Y = k | \mathbf{z}_i)=
  \mathrm{softmax}(
    \mathbf{W}
    \cdot
    \mathbf{z}_i + \textbf{b}
  ).
$
The K+1 classes include K inlier classes 
and one no-object class.
Note that 
$\mathbf{z}_i$
denotes the vector of mask-wide 
pre-logit activations
of the i-th mask.
The mask decoder projects pre-logits $\mathbf{z}_i$
into logits by applying the learned matrix $\mathbf{W}$.
The binary masks $\mathbf{m} = \sigma (\text{conv}_{1 \times 1} (\mathbf{E}, \mathbf{q}))$ are obtained
by scoring per-pixel features $\mathbf{E}$ 
with the mask embeddings $\mathbf{q}$. 
The sigmoid activation interprets 
each element of the obtained tensor 
as a probabilistic assignment of the particular pixel 
into the corresponding mask. 
Semantic segmentation can be carried out 
by classifying each pixel 
according to a weighted ensemble 
of per-mask classifiers $P(y|\mathbf{z})$, 
where the weights correspond 
to dense mask assignments $\mathbf{m}$:
\begin{equation}
   \hat{y}[r,c] = \underset{k=1,\dots,K}{\text{argmax}} \sum_i^N \mathbf{m}_i [r,c] \, P (Y = k | \mathbf{z}_i).
   \label{eq:m2f-classification}
\end{equation}

Default mask-level posterior 
already includes K+1 classes
that correspond to K inlier classes 
and one no-object class. 
We implement our method 
by introducing an additional class 
to learn the negative objectness,
which brings us to K+2 classes in total. 
To be consistent with the image-wide setup, 
in addition to the K inlier classes, 
we place the outlier class at index K+1 
and the no-object class at index K+2.
We expose the segmentation model to negative data by training on mixed-content images \cite{chan2021entropy}. 
Closed-set recognition can still be carried out 
by considering only the K inlier logits 
(\ref{eq:m2f-classification}).

We detect anomalies on the mask-level by applying  
UNO to the mask-wide pre-logits $\textbf{z}_i$ of the i-th mask.
We define the per-pixel outlier score at spatial positions $r$ and $c$ as a sum of mask-level outlier scores weighted with with dense probabilistic mask assignments \cite{grcic2023advantages}:
\begin{align}
   \textbf{s}_{\text{UNO}}^{\text{M2F}}[r,c] = 
   \sum_{i=1}^N \mathbf{m}_i[r,c] \cdot
      \textbf{s}_{\text{UNO}}(\textbf{z}_i)
      \;. 
\end{align}

\section{Experimental setup}
\subsection{Benchmarks and datasets}
We evaluate UNO on 
pixel-level outlier detection benchmarks  
Fishyscapes \cite{blum2021fishyscapes} and 
SMIYC \cite{chan2021segmentmeifyoucan} and the image-wide
OpenOOD \cite{zhang2023openood} benchmark. 

\noindent
\textbf{Pixel-level benchmarks.}
Fishyscapes \cite{blum2021fishyscapes} contains datasets with real (FS Lost\&Found) and synthetic (FS Static) outliers.
SMIYC \cite{chan2021segmentmeifyoucan} has two dominant tracks which group anomalies according to size.
AnomalyTrack focuses on the detection of large anomalies on the traffic scenes while ObstacleTrack focuses on the detection of small obstacles on the road.
Additionaly, we validate on the RoadAnomaly \cite{lis2019detecting} dataset which is an early version of the AnomalyTrack dataset. 

\noindent
\textbf{Image-level benchmarks.}
OpenOOD \cite{zhang2023openood} proposed a unified benchmark for image-wide OOD detection with large-scale datasets. 
The test outliers are divided into two groups (Near-OOD and Far-OOD) based on semantical similarity to the inlier classes or observed empirical difficulty.
The far-OOD group consists of outliers that are semantically far from the inliers (numerical digits, textural patterns, or scene imagery).
The near-OOD group consists of outliers that are semantically similar to the inliers as they all include specific objects.
The OpenOOD-CIFAR-10 setup uses the official CIFAR-10 \cite{krizhevsky2009learning} splits as ID train, val and test subsets. The negative dataset corresponds to a subset of Tiny ImageNet (TIN) \cite{le2015tiny}. The near-OOD datasets include 
CIFAR-100 \cite{krizhevsky2009cifar} and a subset of Tiny ImageNet (TIN) \cite{le2015tiny} that does not overlap with CIFAR-10 and the negative dataset. 
The far-OOD group consists of MNIST \cite{deng2012mnist}, SVHN \cite{yuval2011reading}, Textures \cite{cimpoi2014describing}, and Places365 \cite{zhou2017places} without images that are related with any of the ID classes. 
The large-scale OpenOOD ImageNet-200 benchmark considers a subset of 200 classes from ImageNet-1K\cite{deng09cvpr} as the inlier training dataset. The remaining 800 classes are used as the negative dataset. The near-OOD group consists of SSB-hard \cite{vaze2021open} and NINCO \cite{bitterwolf2023or} while the far-OOD group includes iNaturalist \cite{van2018inaturalist}, Textures \cite{cimpoi2014describing}, and OpenImage-O \cite{wang2022vim}. 

\subsection{Evaluation metrics}
We use standard evaluation metrics: area under the precision-recall curve (AP), area under the receiver operating curve (AUROC or AUC), and false positive rate at $95\%$ true positive rate ($\text{FPR}_{95}$). We validate in-distribution performance with accuracy and mIoU. Note that we omit the AUROC metric in the pixel-level experiments since all methods achieve high AUROC within variance. 

\subsection{Implementation details}
\noindent
\textbf{Segmentation of road scenes.}
Our pixel-level experiments build upon a
Mask2Former model \cite{cheng2022masked} with an ImageNet-initialized SWIN-L \cite{liu2021swin} backbone.
We pre-train the Mask2Former in closed-set setup for 115K iterations on Cityscapes \cite{cordts2016cityscapes} and Mapillary Vistas \cite{neuhold2017mapillary} with the Cityscapes taxonomy. 
We use the default hyperparameters \cite{cheng2022masked} and set the batch size to 18. 
We extend the mask-wide classifier to K+2 classes and fine-tune the model on mixed-content scenes with either real or synthetic negatives for 2K iterations.
We use the zero-initialization for the added weights.
When training with real negative data, we assemble mixed-content images by pasting 
three semantically different instances sampled from ADE20K \cite{zhou2019semantic} after resizing to the range of [96, 512]. 
In the case of training with synthetic negatives, 
we jointly fine tune the K+2-way segmentation model and a flow by adding the loss defined in Equation 6 in the main manuscript to the standard optimization process \cite{cheng2022masked}. 
We use the DenseFlow-25-6 \cite{grcic21neurips} that we pretrain on Mapillary Vistas for 300 epochs.
We generate rectangular patches with spatial dimensions in the range of [48, 512] by sampling the jointly trained flow.
We set the loss modulation parameter from Equation 6 to 0.03. 
We find that training our segmentation model with negatives from scratch, besides from beeing computationally expensive, leads to overfitting to negatives. Thus, we only finetune the model trained in closed-set manner with both real and synthetic negatives.
Additionally, in this case we do not observe feature collapse when utilizing the joint loss (Equation 6 in the main manuscript) since we only jointly train for a small number of iterations. 
The closed-set training of Mask2Former lasts 48 hours, while the fine-tuning stage takes only 30 minutes on three A6000 GPUs. \\

\noindent
\textbf{Image classification}
Our image-wide experiments follow the official training setup \cite{zhang2023openood}. 
When training with real negatives, we train the ResNet-18 \cite{he2016deep} backbone with a K+1-way classification layer for 100 epochs from random initialization. 
We use the SGD optimizer with a momentum of 0.9 and a learning rate of 0.1 with cosine annealing decay schedule. We apply the weight decay of 0.0005. We set the batch size to 128 for CIFAR-10 and 256 for ImageNet-200.
Each minibatch contains the equal ratio of all K+1 classes. Specifically, for CIFAR-10 the minibatch contains 117 inlier images and 11 negative images, and 255 inliers and only one negative sample for the ImageNet-200. 
When training with synthetic negatives we follow the two step procedure as explained in Section 4 of the main manuscript. 
We use the DenseFlow-25-6 \cite{grcic21arxiv} to generate synthetic samples. We pretrain the flow on inlier images, e.g. CIFAR-10 or ImageNet-200, for 300 epoch following the hyperparameters from \cite{grcic21arxiv}.
In the first step, we jointly train the flow pretrained on inlier images and a randomly initialized K-way classifier with the ResNet-18 backbone according to Equation 6 from the main manuscript. We train for 100 epochs and use the same hyperparameters as described above.
In the second step, we freeze the flow and add the K+1-th logit to the classifier and finetune for 20 epochs.
We set the learning rate for the backbone to 0.0001 and 0.01 for the final fully connected layer.
We set the loss modulation parameter from Equation 6 to 0.03. 


\section{Additional results}

We provide 
a discussion on the choice of negative data, 
an alternative implementation of negative objectness and the full results on the  OpenOOD \cite{zhang2023openood} benchmark.

\subsection{Impact of synthetic negatives}
Table~\ref{tab:negatives-abliation} shows the performance of UNO depending on the source of negative training data. 
We experiment with random crops from inlier scenes, synthetic negatives generated by a jointly trained normalizing flow, and random instances from the ADE datasets.
Training on synthetic negatives is more beneficial than training on inlier even though the flow was jointly trained only for a small number of iterations.
For instance, training on inlier crops yields a high FPR\textsubscript{95} on Fishyscapes Lost\&Found.
Contrary, synthetic negatives yield low FPR\textsubscript{95} on all three validations sets.
Still, there is a performance gap between models that are trained with and without real negative data.

\begin{table}[ht]
\setlength{\tabcolsep}{4pt}
\centering
\small
\footnotesize
  \caption{Performance of UNO for different training negatives in per-pixel outlier segmentation.}
  \vskip 0.15in
  \label{tab:negatives-abliation}
  \begin{tabular}{lcccccc}
     Training & \multicolumn{2}{c}{FS L\&F} & \multicolumn{2}{c}{FS Static} & \multicolumn{2}{c}{RoadAnomaly}\\
    negatives & AP & $\text{FPR}_{95}$ & AP &  $\text{FPR}_{95}$ & AP & $\text{FPR}_{95}$ \\
    \midrule
    Inlier crops & 67.2  & 62.5 & 79.2 & 0.9 & 86.5 & 7.6 \\
    Synthetic negatives & 74.5  & 6.9 & 96.9  & 0.1 & 82.4 & 9.2 \\
    ADE20k negatives & 81.8  & 1.3 & 98.0  & 0.04 & 88.5 & 7.4 \\
\end{tabular}
\end{table}

\subsection{Alternative implementation of the outlier posterior}
The outlier posterior $P(y_\text{NO}|\mathbf{z})$ can alternatively be modeled with an additional out-of-distribution head \cite{bevandic22ivc}. 
Then, we introduce an additional binary cross-entropy loss term to train the out-of-distribution head that discriminates inliers and outliers.
This way the closed set classifier ends up with K classes and is not affected by the negative data.
When applied to the mask-wide recognition architecture, the OOD head has 3 outputs 1) inlier, 2) outlier and 3) no-object. 
Table~\ref{tab:ood-head} shows the comparison of the K+2-way classifier proposed in the main paper with a K+1-way classifier and a 3-way OOD head and ablation of UNO components in the pixel-wise outlier detection setup.
Our original UNO formulation built atop K+2-way classifier consistently outperforms alternative formulations across all datasets and metrics. 
\begin{table}[ht]
\centering
\footnotesize
  \caption{Comparison of the K+2-way classifier with the OOD head atop of Mask2Former architecture on Fishyscapes val and RoadAnomaly.}
  \label{tab:ood-head}
  \begin{tabular}{llcccccc}
     && \multicolumn{2}{c}{FS L\&F} & \multicolumn{2}{c}{FS Static} & \multicolumn{2}{c}{RoadAnomaly}\\
    Method & Score & AP & $\text{FPR}_{95}$ & AP & $\text{FPR}_{95}$ & AP & $\text{FPR}_{95}$ \\
    \midrule
    K+2-way classifier & $\text{s}_{\text{UNO}}$ & 81.8  & 1.3 & 98.0 & 0.0	& 88.5 & 7.4 \\
    K-way classifier \& OOD head & $\text{s}_{\text{UNO}}$ & 81.4 & 5.3 &  89.0 & 0.3 & 80.6 & 10.6 \\
    K-way classifier \& OOD head & $\text{s}_{\text{Unc}}$  & 77.8 & 2.2 & 86.4 & 1.6 & 76.1 & 8.4 \\ 
    K-way classifier \& OOD head & $\text{s}_\text{NO}$  & 79.9 & 7.3  &  88.3 & 0.3  & 74.5 & 25.2  \\

\end{tabular}
\end{table}

Table~\ref{tab:cifar-TH} presents a similar analysis in the image-wide setting using the OpenOOD CIFAR-10 setup.
Again, the K+1-way classifier combined with our UNO score outperforms the K-way classifier and a binary OOD head.
Still, our UNO score works well even with the alternative formulation.
\begin{table}[ht]
\setlength{\tabcolsep}{4pt}
    \centering
    \small
    \footnotesize
    \caption{Comparison of the K+1-way classifier with the binary OOD head atop of ResNet-18 on OpenOOD CIFAR-10. We use UNO as the outlier score.}
    \vskip 0.15in
    \begin{tabular}{lcccc}
         \multirow{2}{*}{Method} & \multicolumn{2}{c}{Near-OOD} & \multicolumn{2}{c}{Far-OOD} \\
          & AUC & $\text{FPR}_{95}$ & AUC & $\text{FPR}_{95}$\\
         \hline
        K+1-way classifier & 95.00 & 18.75 & 97.95 & 9.30 \\	
        K-way classifier \& OOD head & 94.23 & 19.31 & 97.76 & 11.12 \\	
    \end{tabular}
    
    \label{tab:cifar-TH}
\end{table}

\subsection{Full results on OpenOOD}
Tables~\ref{tab:cifar-10-full} and~\ref{tab:imagenet-200-full} provide the extended results on the OpenOOD \cite{zhang2023openood} benchmark. We compare with post-hoc methods (upper section) and training methods without (middle section) and with the use of real negative data (bottom section).
All results are averaged over three runs with variances in subscripts.
\begin{table}
\centering
  \caption{OOD detection performance on the OpenOOD benchmark, the CIFAR-10 dataset.
  }
  \vskip 0.15in
  \footnotesize
  \label{tab:cifar-10-full}
  \begin{tabular}{l
   c@{\ }c@{\ }c@{\quad}c@{\ }c@{\ }c@{\quad}c@{\ }
  }
    \multirow{2}{*}{Method}
    &  \multicolumn{3}{c}{Near-OOD} & \multicolumn{3}{c}{Far-OOD}\\

     & AUC & FPR\textsubscript{95}  & AP & AUC & FPR\textsubscript{95} & AP & Acc.\\

    \midrule
    OpenMax \cite{bendale2016cvpr}
        & 87.62$_{(\pm0.29)}$ & 43.62$_{(\pm2.27)}$ & 80.60$_{(\pm0.29)}$
        & 89.62$_{(\pm0.19)}$ & 29.69$_{(\pm1.21)}$ & 90.19$_{(\pm0.41)}$
        & 95.06$_{(\pm0.30)}$ \\
    MSP \cite{hendrycks17iclr}
        & 88.03$_{(\pm0.25)}$ & 48.17$_{(\pm3.92)}$ & 85.43$_{(\pm0.36)}$
        & 90.73$_{(\pm0.43)}$ & 31.72$_{(\pm1.84)}$ & 93.27$_{(\pm0.14)}$
        & 95.06$_{(\pm0.30)}$ \\
    TempScale \cite{guo2017calibration}
        & 88.09$_{(\pm0.31)}$ & 50.96$_{(\pm4.32)}$ & 86.11$_{(\pm0.33)}$
        & 90.97$_{(\pm0.52)}$ & 33.48$_{(\pm2.39)}$ & 93.68$_{(\pm0.11)}$
        & 95.06$_{(\pm0.30)}$ \\
    ODIN \cite{liang2017enhancing}
        & 82.87$_{(\pm1.85)}$ & 76.19$_{(\pm6.08)}$ & 83.03$_{(\pm1.15)}$
        & 87.96$_{(\pm0.61)}$ & 57.62$_{(\pm4.24)}$ & 93.14$_{(\pm0.44)}$
        & 95.06$_{(\pm0.30)}$ \\
    MDS \cite{Lee2018neurips}
        & 84.20$_{(\pm2.40)}$ & 49.90$_{(\pm3.98)}$ & 79.88$_{(\pm3.18)}$
        & 89.72$_{(\pm1.36)}$ & 32.22$_{(\pm3.40)}$ & 93.81$_{(\pm0.74)}$
        & 95.06$_{(\pm0.30)}$ \\
    MDSEns \cite{Lee2018neurips}
        & 60.43$_{(\pm0.26)}$ & 92.26$_{(\pm0.20)}$ & 59.94$_{(\pm0.19)}$
        & 73.90$_{(\pm0.27)}$ & 61.47$_{(\pm0.48)}$ & 83.37$_{(\pm0.04)}$
        & 95.06$_{(\pm0.30)}$ \\
    RMDS \cite{ren2021arxiv}
        & 89.80$_{(\pm0.28)}$ & 38.89$_{(\pm2.39)}$ & 87.52$_{(\pm0.29)}$
        & 92.20$_{(\pm0.21)}$ & 25.35$_{(\pm0.73)}$ & 94.21$_{(\pm0.10)}$
        & 95.06$_{(\pm0.30)}$ \\
    Gram \cite{sastry2020icml}
        & 58.66$_{(\pm4.83)}$ & 90.87$_{(\pm1.91)}$ & 57.57$_{(\pm5.09)}$
        & 71.73$_{(\pm3.20)}$ & 72.34$_{(\pm6.73)}$ & 82.89$_{(\pm3.14)}$
        & 95.06$_{(\pm0.30)}$ \\
    EBO \cite{liu2020neurips}
        & 87.58$_{(\pm0.46)}$ & 61.34$_{(\pm4.63)}$ & 87.04$_{(\pm0.27)}$
        & 91.21$_{(\pm0.92)}$ & 41.69$_{(\pm5.32)}$ & 94.31$_{(\pm0.09)}$
        & 95.06$_{(\pm0.30)}$ \\
    OpenGAN \cite{kong2021iccv}
        & 53.71$_{(\pm7.68)}$ & 94.48$_{(\pm4.01)}$ & 53.35$_{(\pm5.22)}$
        & 54.61$_{(\pm15.51)}$ & 83.52$_{(\pm11.63)}$ & 73.34$_{(\pm8.49)}$
        & 95.06$_{(\pm0.30)}$ \\
    GradNorm \cite{huang2021neurips}
        & 54.90$_{(\pm0.98)}$ & 94.72$_{(\pm0.82)}$ & 57.95$_{(\pm1.98)}$
        & 57.55$_{(\pm3.22)}$ & 91.90$_{(\pm2.23)}$ & 76.75$_{(\pm1.95)}$
        & 95.06$_{(\pm0.30)}$ \\
    ReAct \cite{sun2021neurips}
        & 87.11$_{(\pm0.61)}$ & 63.56$_{(\pm7.33)}$ & 86.65$_{(\pm0.19)}$
        & 90.42$_{(\pm1.41)}$ & 44.90$_{(\pm8.37)}$ & 93.99$_{(\pm0.45)}$
        & 95.06$_{(\pm0.30)}$ \\
    MLS \cite{hendrycks2022icml}
        & 87.52$_{(\pm0.47)}$ & 61.32$_{(\pm4.62)}$ & 86.88$_{(\pm0.29)}$
        & 91.10$_{(\pm0.89)}$ & 41.68$_{(\pm5.27)}$ & 94.21$_{(\pm0.06)}$
        & 95.06$_{(\pm0.30)}$ \\
    KLM \cite{hendrycks2022icml}
        & 79.19$_{(\pm0.80)}$ & 87.86$_{(\pm6.37)}$ & 80.37$_{(\pm0.46)}$
        & 82.68$_{(\pm0.21)}$ & 78.31$_{(\pm4.84)}$ & 90.57$_{(\pm0.25)}$
        & 95.06$_{(\pm0.30)}$ \\
    VIM \cite{wang2022vim}
        & 88.68$_{(\pm0.28)}$ & 44.84$_{(\pm2.31)}$ & 86.32$_{(\pm0.39)}$
        & 93.48$_{(\pm0.24)}$ & 25.05$_{(\pm0.52)}$ & 96.27$_{(\pm0.24)}$
        & 95.06$_{(\pm0.30)}$ \\
    KNN \cite{sun2022icml}
        & 90.64$_{(\pm0.20)}$ & 34.01$_{(\pm0.38)}$ & 88.50$_{(\pm0.35)}$
        & 92.96$_{(\pm0.14)}$ & 24.27$_{(\pm0.40)}$ & 94.93$_{(\pm0.07)}$ 
        & 95.06$_{(\pm0.30)}$ \\
    DICE \cite{sun2022eccv}
        & 78.34$_{(\pm0.79)}$ & 70.04$_{(\pm7.64)}$ & 74.80$_{(\pm2.33)}$
        & 84.23$_{(\pm1.89)}$ & 51.76$_{(\pm4.42)}$ & 89.06$_{(\pm1.72)}$
        & 95.06$_{(\pm0.30)}$ \\
    RankFeat \cite{song2022neurips}
        & 79.46$_{(\pm2.52)}$ & 60.88$_{(\pm4.60)}$ & 74.46$_{(\pm3.08)}$ & 75.87$_{(\pm5.06)}$ & 57.44$_{(\pm7.99)}$ & 81.27$_{(\pm3.67)}$ & 95.06$_{(\pm0.30)}$\\
    ASH \cite{djurisic2022extremely}
        & 75.27$_{(\pm1.04)}$ & 86.78$_{(\pm1.82)}$ & 77.24$_{(\pm1.26)}$ & 78.49$_{(\pm2.58)}$ & 79.03$_{(\pm4.22)}$ & 88.33$_{(\pm1.39)}$ & 95.06$_{(\pm0.30)}$\\
    SHE \cite{zhang2023iclr}
        & 81.54$_{(\pm0.51)}$ & 79.65$_{(\pm3.47)}$ & 82.04$_{(\pm0.51)}$ & 85.32$_{(\pm1.43)}$ & 66.48$_{(\pm5.98)}$ & 91.26$_{(\pm0.04)}$ & 95.06$_{(\pm0.30)}$\\

    \midrule
    ConfBranch \cite{devries2018arxiv}
        & 89.84$_{(\pm0.24)}$ & 31.28$_{(\pm0.66)}$ & 85.50$_{(\pm0.30)}$ 
        & 92.85$_{(\pm0.29)}$ & 94.88$_{(\pm0.05)}$ & 93.48$_{(\pm0.39)}$
        & 94.88$_{(\pm0.05)}$\\
    RotPred \cite{hendrycks2019using}
        & 92.68$_{(\pm0.27)}$ & 28.14$_{(\pm1.68)}$ & 90.47$_{(\pm0.35)}$ 
        & 96.62$_{(\pm0.18)}$ & 12.23$_{(\pm0.33)}$ & 97.54$_{(\pm0.13)}$ 
        & 95.35$_{(\pm0.52)}$\\
    G-ODIN \cite{hsu2020cvpr}
        & 89.12$_{(\pm0.57)}$ & 45.54$_{(\pm2.52)}$ & 88.25$_{(\pm0.49)}$ 
        & 95.51$_{(\pm0.31)}$ & 21.45$_{(\pm1.91)}$ & 97.35$_{(\pm0.34)}$ 
        & 94.70$_{(\pm0.25)}$\\
    CSI \cite{tack2020neurips}
        & 89.51$_{(\pm0.19)}$ & 33.66$_{(\pm0.64)}$ & 86.37$_{(\pm0.25)}$ 
        & 92.00$_{(\pm0.30)}$ & 26.42$_{(\pm0.29)}$ & 93.90$_{(\pm0.33)}$ 
        & 91.16$_{(\pm0.14)}$ \\
    ARPL \cite{chen2022pami}
        & 87.44$_{(\pm0.15)}$ & 40.33$_{(\pm0.70)}$ & 82.96$_{(\pm0.33)}$ 
        & 89.31$_{(\pm0.32)}$ & 32.39$_{(\pm0.74)}$ & 91.41$_{(\pm0.09)}$ 
        & 93.66$_{(\pm0.11)}$\\
    MOS \cite{huang2021cvpr}
        & 71.45$_{(\pm3.09)}$ & 78.72$_{(\pm5.86)}$ & 72.41$_{(\pm3.05)}$ 
        & 76.41$_{(\pm5.93)}$ & 62.90$_{(\pm6.62)}$ & 85.24$_{(\pm2.92)}$ 
        & 94.83$_{(\pm0.37)}$\\
    VOS \cite{du2022iclr}
        & 87.70$_{(\pm0.48)}$ & 57.03$_{(\pm1.92)}$ & 86.57$_{(\pm0.73)}$ 
        & 90.83$_{(\pm0.92)}$ & 40.43$_{(\pm4.53)}$ & 93.95$_{(\pm0.56)}$
        & 94.31$_{(\pm0.64)}$\\
    LogitNorm \cite{wei2022icml}
        & 92.33$_{(\pm0.08)}$ & 29.34$_{(\pm0.81)}$ & 90.62$_{(\pm0.09)}$ 
        & 96.74$_{(\pm0.06)}$ & 13.81$_{(\pm0.20)}$ & 97.29$_{(\pm0.21)}$ 
        & 94.30$_{(\pm0.25)}$\\
    CIDER \cite{ming2023iclr}
        & 90.71$_{(\pm0.16)}$ & 32.11$_{(\pm0.94)}$ & 87.97$_{(\pm0.24)}$ 
        & 94.71$_{(\pm0.36)}$ & 20.72$_{(\pm0.85)}$ & 96.19$_{(\pm0.19)}$ 
        & - \\
    NPOS \cite{tao2023iclr}
        & 89.78$_{(\pm0.33)}$ & 32.64$_{(\pm0.70)}$ & 86.36$_{(\pm0.68)}$ 
        & 94.07$_{(\pm0.49)}$ & 20.59$_{(\pm0.69)}$ & 96.20$_{(\pm0.43)}$ 
        & - \\
    UNO (ours) &
        91.34$_{(\pm0.33)}$&31.78$_{(\pm0.82)}$& 87.39$_{(\pm0.35)}$ &
        92.55$_{(\pm0.25)}$&20.54$_{(\pm0.87)}$& 93.96$_{(\pm0.25)}$ &
        95.20$_{(\pm0.25)}$\\
    \midrule 
    MixOE \cite{zhang2023mixture}
         & 88.73$_{(\pm0.82)}$ & 51.45$_{(\pm7.78)}$ & 94.25$_{(\pm0.17)}$ 
         & 91.93$_{(\pm0.69)}$ & 33.84$_{(\pm4.77)}$ & 98.00$_{(\pm0.02)}$ 
         & 94.55$_{(\pm0.32)}$\\
    MCD \cite{yu2019unsupervised}
         & 91.03$_{(\pm0.12)}$ & 30.17$_{(\pm0.06)}$ & 87.73$_{(\pm0.36)}$ 
         & 91.00$_{(\pm1.10)}$ & 32.03$_{(\pm4.21)}$ & 94.45$_{(\pm0.85)}$ 
         & 94.95$_{(\pm0.04)}$ \\
    UDG \cite{yang2021semantically}
        & 89.91$_{(\pm0.25)}$ & 35.34$_{(\pm0.95)}$ & 86.89$_{(\pm0.84)}$ 
        & 94.06$_{(\pm0.90)}$ & 20.35$_{(\pm2.41)}$ & 95.23$_{(\pm0.89)}$ 
        & 92.36$_{(\pm0.84)}$\\
    OE \cite{hendrycks17iclr} 
        & 94.82$_{(\pm0.21)}$ & 19.84$_{(\pm0.95)}$ & 87.39$_{(\pm0.60)}$ 
        & 96.00$_{(\pm0.13)}$ & 13.13$_{(\pm0.53)}$ & 95.03$_{(\pm0.12)}$ 
        & 94.63$_{(\pm0.26)}$\\
    UNO (ours)
        & 94.87$_{(\pm0.07)}$ & 9.33$_{(\pm0.50)}$ & 94.49$_{(\pm0.13)}$
        & 97.63$_{(\pm0.72)}$ & 9.38$_{(\pm2.65)}$ & 99.10$_{(\pm0.25)}$
        & 94.88$_{(\pm0.19)}$ \\

  \end{tabular}
\end{table}

\begin{table}
\centering
  \caption{OOD detection performace on the OpenOOD benchmark, the Imagenet-200 dataset.
  }
  \vskip 0.15in
  \footnotesize
  \label{tab:imagenet-200-full}
    \begin{tabular}{l
   c@{\ }c@{\ }c@{\quad}c@{\ }c@{\ }c@{\quad}c@{\ }
  }
    \multirow{2}{*}{Method}
    &  \multicolumn{3}{c}{Near-OOD} & \multicolumn{3}{c}{Far-OOD}\\

     & AUC & FPR\textsubscript{95}  & AP & AUC & FPR\textsubscript{95} & AP & Acc.\\

    \midrule
OpenMax \cite{bendale2016cvpr}
& 80.27$_{(\pm0.10)}$ & 63.48$_{(\pm0.25)}$ & 81.42$_{(\pm0.19)}$ & 90.20$_{(\pm0.17)}$ & 33.12$_{(\pm0.66)}$ & 85.13$_{(\pm0.47)}$ & 86.37$_{(\pm0.08)}$\\
MSP \cite{hendrycks17iclr}
& 83.34$_{(\pm0.06)}$ & 54.82$_{(\pm0.35)}$ & 85.95$_{(\pm0.05)}$ & 90.13$_{(\pm0.09)}$ & 35.43$_{(\pm0.38)}$ & 88.71$_{(\pm0.14)}$ & 86.37$_{(\pm0.08)}$\\
TempScale \cite{guo2017calibration}
& 83.69$_{(\pm0.04)}$ & 54.82$_{(\pm0.23)}$ & 86.29$_{(\pm0.02)}$ & 90.82$_{(\pm0.09)}$ & 34.00$_{(\pm0.37)}$ & 89.49$_{(\pm0.15)}$ & 86.37$_{(\pm0.08)}$\\
ODIN \cite{liang2017enhancing}
& 80.27$_{(\pm0.08)}$ & 66.76$_{(\pm0.26)}$ & 85.02$_{(\pm0.03)}$ & 91.71$_{(\pm0.19)}$ & 34.23$_{(\pm1.05)}$ & 91.29$_{(\pm0.15)}$ & 86.37$_{(\pm0.08)}$\\
MDS \cite{Lee2018neurips}
& 61.93$_{(\pm0.51)}$ & 79.11$_{(\pm0.31)}$ & 67.68$_{(\pm0.42)}$ & 74.72$_{(\pm0.26)}$ & 61.66$_{(\pm0.27)}$ & 70.80$_{(\pm0.61)}$ & 86.37$_{(\pm0.08)}$\\
MDSEns \cite{Lee2018neurips}
& 54.32$_{(\pm0.24)}$ & 91.75$_{(\pm0.10)}$ & 64.81$_{(\pm0.24)}$ & 69.27$_{(\pm0.57)}$ & 80.96$_{(\pm0.38)}$ & 69.62$_{(\pm0.52)}$ & 86.37$_{(\pm0.08)}$\\
RMDS \cite{ren2021arxiv}
& 82.57$_{(\pm0.25)}$ & 54.02$_{(\pm0.58)}$ & 83.07$_{(\pm0.45)}$ & 88.06$_{(\pm0.34)}$ & 32.45$_{(\pm0.79)}$ & 82.71$_{(\pm0.78)}$ & 86.37$_{(\pm0.08)}$\\
Gram \cite{sastry2020icml}
& 67.67$_{(\pm1.07)}$ & 86.40$_{(\pm1.21)}$ & 75.63$_{(\pm0.78)}$ & 71.19$_{(\pm0.24)}$ & 84.36$_{(\pm0.78)}$ & 72.75$_{(\pm0.25)}$ & 86.37$_{(\pm0.08)}$\\
EBO \cite{liu2020neurips}
& 82.50$_{(\pm0.05)}$ & 60.24$_{(\pm0.57)}$ & 85.48$_{(\pm0.07)}$ & 90.86$_{(\pm0.21)}$ & 34.86$_{(\pm1.30)}$ & 89.85$_{(\pm0.21)}$ & 86.37$_{(\pm0.08)}$\\
OpenGAN \cite{kong2021iccv}
& 59.79$_{(\pm3.39)}$ & 84.15$_{(\pm3.85)}$ & 66.85$_{(\pm2.79)}$ & 73.15$_{(\pm4.07)}$ & 64.16$_{(\pm9.33)}$ & 66.62$_{(\pm3.69)}$ & 86.37$_{(\pm0.08)}$\\
GradNorm \cite{huang2021neurips}
& 72.75$_{(\pm0.48)}$ & 82.67$_{(\pm0.30)}$ & 80.19$_{(\pm0.68)}$ & 84.26$_{(\pm0.87)}$ & 66.45$_{(\pm0.22)}$ & 86.54$_{(\pm0.92)}$ & 86.37$_{(\pm0.08)}$\\
ReAct \cite{sun2021neurips}
& 81.87$_{(\pm0.98)}$ & 62.49$_{(\pm2.19)}$ & 85.38$_{(\pm0.34)}$ & 92.31$_{(\pm0.56)}$ & 28.50$_{(\pm0.95)}$ & 91.31$_{(\pm0.80)}$ & 86.37$_{(\pm0.08)}$\\
MLS \cite{hendrycks2022icml}
& 82.90$_{(\pm0.04)}$ & 59.76$_{(\pm0.59)}$ & 85.96$_{(\pm0.07)}$ & 91.11$_{(\pm0.19)}$ & 34.03$_{(\pm1.21)}$ & 90.10$_{(\pm0.21)}$ & 86.37$_{(\pm0.08)}$\\
KLM \cite{hendrycks2022icml}
& 80.76$_{(\pm0.08)}$ & 70.26$_{(\pm0.64)}$ & 83.41$_{(\pm0.23)}$ & 88.53$_{(\pm0.11)}$ & 40.90$_{(\pm1.08)}$ & 84.22$_{(\pm0.47)}$ & 86.37$_{(\pm0.08)}$\\
VIM \cite{wang2022vim}
& 78.68$_{(\pm0.24)}$ & 59.19$_{(\pm0.71)}$ & 81.61$_{(\pm0.29)}$ & 91.26$_{(\pm0.19)}$ & 27.20$_{(\pm0.30)}$ & 90.01$_{(\pm0.35)}$ & 86.37$_{(\pm0.08)}$\\
KNN \cite{sun2022icml}
& 81.57$_{(\pm0.17)}$ & 60.18$_{(\pm0.52)}$ & 85.72$_{(\pm0.17)}$ & 93.16$_{(\pm0.22)}$ & 27.27$_{(\pm0.75)}$ & 93.48$_{(\pm0.15)}$ & 86.37$_{(\pm0.08)}$\\
DICE \cite{sun2022eccv}
& 81.78$_{(\pm0.14)}$ & 61.88$_{(\pm0.67)}$ & 85.37$_{(\pm0.13)}$ & 90.80$_{(\pm0.31)}$ & 36.51$_{(\pm1.18)}$ & 90.55$_{(\pm0.29)}$ & 86.37$_{(\pm0.08)}$\\
RankFeat \cite{song2022neurips}
& 56.92$_{(\pm1.59)}$ & 92.06$_{(\pm0.23)}$ & 66.17$_{(\pm1.63)}$ & 38.22$_{(\pm3.85)}$ & 97.72$_{(\pm0.75)}$ & 45.25$_{(\pm2.81)}$ & 86.37$_{(\pm0.08)}$\\
ASH \cite{djurisic2022extremely}
& 82.38$_{(\pm0.19)}$ & 64.89$_{(\pm0.90)}$ & 87.03$_{(\pm0.06)}$ & 93.90$_{(\pm0.27)}$ & 27.29$_{(\pm1.12)}$ & 94.15$_{(\pm0.32)}$ & 86.37$_{(\pm0.08)}$\\
SHE \cite{zhang2023iclr}
& 80.18$_{(\pm0.25)}$ & 66.80$_{(\pm0.74)}$ & 84.20$_{(\pm0.28)}$ & 89.81$_{(\pm0.61)}$ & 42.17$_{(\pm1.24)}$ & 90.05$_{(\pm0.62)}$ & 86.37$_{(\pm0.08)}$\\

\midrule
ConfBranch \cite{devries2018arxiv}
& 79.10$_{(\pm0.24)}$ & 61.44$_{(\pm0.34)}$ & 82.11$_{(\pm0.30)}$ & 90.43$_{(\pm0.18)}$ & 34.75$_{(\pm0.63)}$ & 88.67$_{(\pm0.27)}$ & 85.92$_{(\pm0.07)}$\\
RotPred \cite{hendrycks2019using}
& 81.59$_{(\pm0.20)}$ & 60.42$_{(\pm0.60)}$ & 84.87$_{(\pm0.19)}$ & 92.56$_{(\pm0.09)}$ & 26.16$_{(\pm0.38)}$ & 90.10$_{(\pm0.08)}$ & 86.37$_{(\pm0.16)}$\\
G-ODIN \cite{hsu2020cvpr}
& 77.28$_{(\pm0.10)}$ & 69.87$_{(\pm0.46)}$ & 82.77$_{(\pm0.16)}$ & 92.33$_{(\pm0.11)}$ & 30.18$_{(\pm0.49)}$ & 92.04$_{(\pm0.10)}$ & 84.56$_{(\pm0.28)}$\\
ARPL \cite{chen2022pami}
& 82.02$_{(\pm0.10)}$ & 55.74$_{(\pm0.70)}$ & 84.35$_{(\pm0.08)}$ & 89.23$_{(\pm0.11)}$ & 36.46$_{(\pm0.08)}$ & 87.63$_{(\pm0.19)}$ & 83.95$_{(\pm0.32)}$\\
MOS \cite{huang2021cvpr}
& 69.84$_{(\pm0.46)}$ & 71.60$_{(\pm0.48)}$ & 73.38$_{(\pm0.56)}$ & 80.46$_{(\pm0.92)}$ & 51.56$_{(\pm0.42)}$ & 72.79$_{(\pm1.43)}$ & 85.60$_{(\pm0.20)}$\\
VOS \cite{du2022iclr}
& 82.51$_{(\pm0.11)}$ & 59.89$_{(\pm0.47)}$ & 85.59$_{(\pm0.07)}$ & 91.00$_{(\pm0.28)}$ & 34.01$_{(\pm0.97)}$ & 90.11$_{(\pm0.30)}$ & 86.23$_{(\pm0.19)}$\\
LogitNorm \cite{wei2022icml}
& 82.66$_{(\pm0.15)}$ & 56.46$_{(\pm0.37)}$ & 86.41$_{(\pm0.08)}$ & 93.04$_{(\pm0.21)}$ & 26.11$_{(\pm0.52)}$ & 92.25$_{(\pm0.32)}$ & 86.04$_{(\pm0.15)}$\\
CIDER \cite{ming2023iclr}
& 80.58$_{(\pm1.75)}$ & 60.10$_{(\pm0.73)}$ & 83.32$_{(\pm1.76)}$ & 90.66$_{(\pm1.68)}$ & 30.17$_{(\pm2.75)}$ & 89.16$_{(\pm2.38)}$ & -\\
NPOS \cite{tao2023iclr}
& 79.40$_{(\pm0.39)}$ & 62.09$_{(\pm0.05)}$ & 84.37$_{(\pm0.35)}$ & 94.49$_{(\pm0.07)}$ & 21.76$_{(\pm0.21)}$ & 94.83$_{(\pm0.07)}$ & -\\

UNO (ours)
& 81.16$_{(\pm0.65)}$ & 61.10$_{(\pm0.93)}$ & 85.31$_{(\pm0.94)}$
& 92.53$_{(\pm0.48)}$ & 32.32$_{(\pm0.19)}$ & 87.24$_{(\pm0.33)}$
& 86.33$_{(\pm0.36)}$\\

\midrule
OE \cite{hendrycks17iclr}
& 84.84$_{(\pm0.16)}$ & 52.30$_{(\pm0.67)}$ & 86.86$_{(\pm0.22)}$ & 89.02$_{(\pm0.18)}$ & 34.17$_{(\pm0.56)}$ & 85.15$_{(\pm0.26)}$ & 85.82$_{(\pm0.21)}$\\
MCD \cite{yu2019unsupervised}
& 83.62$_{(\pm0.09)}$ & 54.71$_{(\pm0.83)}$ & 84.44$_{(\pm0.30)}$ & 88.94$_{(\pm0.10)}$ & 29.93$_{(\pm0.30)}$ & 82.90$_{(\pm0.48)}$ & 86.12$_{(\pm0.17)}$\\
UDG \cite{yang2021semantically}
& 74.30$_{(\pm1.63)}$ & 68.89$_{(\pm1.72)}$ & 78.09$_{(\pm2.02)}$ & 82.09$_{(\pm2.78)}$ & 62.04$_{(\pm5.99)}$ & 81.63$_{(\pm3.29)}$ & 68.11$_{(\pm1.24)}$\\
MixOE \cite{zhang2023mixture}
& 82.62$_{(\pm0.03)}$ & 57.97$_{(\pm0.40)}$ & 84.78$_{(\pm0.05)}$ & 88.27$_{(\pm0.41)}$ & 40.93$_{(\pm0.29)}$ & 86.01$_{(\pm0.60)}$ & 85.71$_{(\pm0.07)}$\\

UNO (ours)
& 85.07$_{(\pm0.78)}$ & 51.71$_{(\pm0.31)}$ & 85.29$_{(\pm0.71)}$
& 89.63$_{(\pm0.46)}$ & 36.79$_{(\pm0.21)}$ & 87.07$_{(\pm0.14)}$
& 86.42$_{(\pm0.32)}$\\

  \end{tabular}
\end{table}


\end{document}